\documentclass[10pt,letterpaper]{article}

\usepackage[pagenumbers]{cvpr} 

\newcommand{\cmark}{\(\checkmark\)}
\newcommand{\xmark}{--}
\newcommand{\pmark}{\(\circ\)}

\definecolor{rowblue}{RGB}{243,244,255}
\definecolor{rowgreen}{RGB}{234,248,236}

\definecolor{cvprblue}{rgb}{0.21,0.49,0.74}
\usepackage[pagebackref,breaklinks,colorlinks,allcolors=cvprblue]{hyperref}

\usepackage{pifont}
\definecolor{darkgreen}{RGB}{0,100,0}
\usepackage{makecell}
\usepackage{tcolorbox}

\usepackage{tabularx}
\usepackage{booktabs}

\def\paperID{*****} 
\def\confName{CVPR}
\def\confYear{2025}

\title{Explain Before You Answer: A Survey on Compositional Visual Reasoning}

\author{
Fucai Ke$^{1}$ \quad
Joy Hsu$^{2}$ \quad
Zhixi Cai$^{1}$ \quad
Zixian Ma$^{3}$ \quad 
Xin Zheng$^{4}$ \quad 
Xindi Wu$^{5}$ \quad \\
Sukai Huang$^{1}$ \quad  
Weiqing Wang$^{1}$ \quad
Pari Delir Haghighi$^{1}$ \quad
Gholamreza Haffari$^{1}$ \quad \\
Ranjay Krishna$^{3,6}$ \quad
Jiajun Wu$^{2}$ \quad
Hamid Rezatofighi$^{1}$ \quad \\ \\
$^{1}$Monash University \quad 
$^{2}$Stanford University \quad 
$^{3}$University of Washington \quad \\
$^{4}$Griffith University \quad 
$^{5}$Princeton University \quad 
$^{6}$Allen Institute for Artificial Intelligence \quad \\
{
\href{https://github.com/pokerme7777/Compositional-Visual-Reasoning-Survey}{Survey Project Page}}
}

\begin{document}
\maketitle

\begin{abstract}
Compositional visual reasoning (CVR) has emerged as a key research frontier in multimodal AI, aiming to endow machines with the human-like ability to decompose visual scenes, ground intermediate concepts, and perform multi-step logical inference. While early surveys focus on monolithic vision-language models or general multimodal reasoning, a dedicated synthesis of the rapidly expanding CVR literature is still lacking. We provide a systematic synthesis of more than 100 studies spanning recent advances in compositional visual reasoning. We first formalize core definitions and describe why compositional approaches offer advantages in cognitive alignment, semantic fidelity, robustness, interpretability, and data efficiency. Next, we trace a five-stage paradigm shift: from prompt-enhanced language-centric pipelines, through tool-enhanced LLMs and tool-enhanced VLMs, to recent chain-of-thought reasoning and unified agentic VLMs, highlighting their architectural designs, strengths, and limitations. 
We summarize 60+ benchmarks and commonly used metrics relevant to CVR. 
Drawing on these analyses, we distill key insights, identify open challenges and outline future directions. By offering a unified taxonomy, historical roadmap, and critical outlook, this survey aims to serve as a structured reference and inspire the next generation of CVR research.

\end{abstract}

\tableofcontents
\section{Introduction}

\begin{figure*}
\begin{center}
  \includegraphics[width=0.9\linewidth]{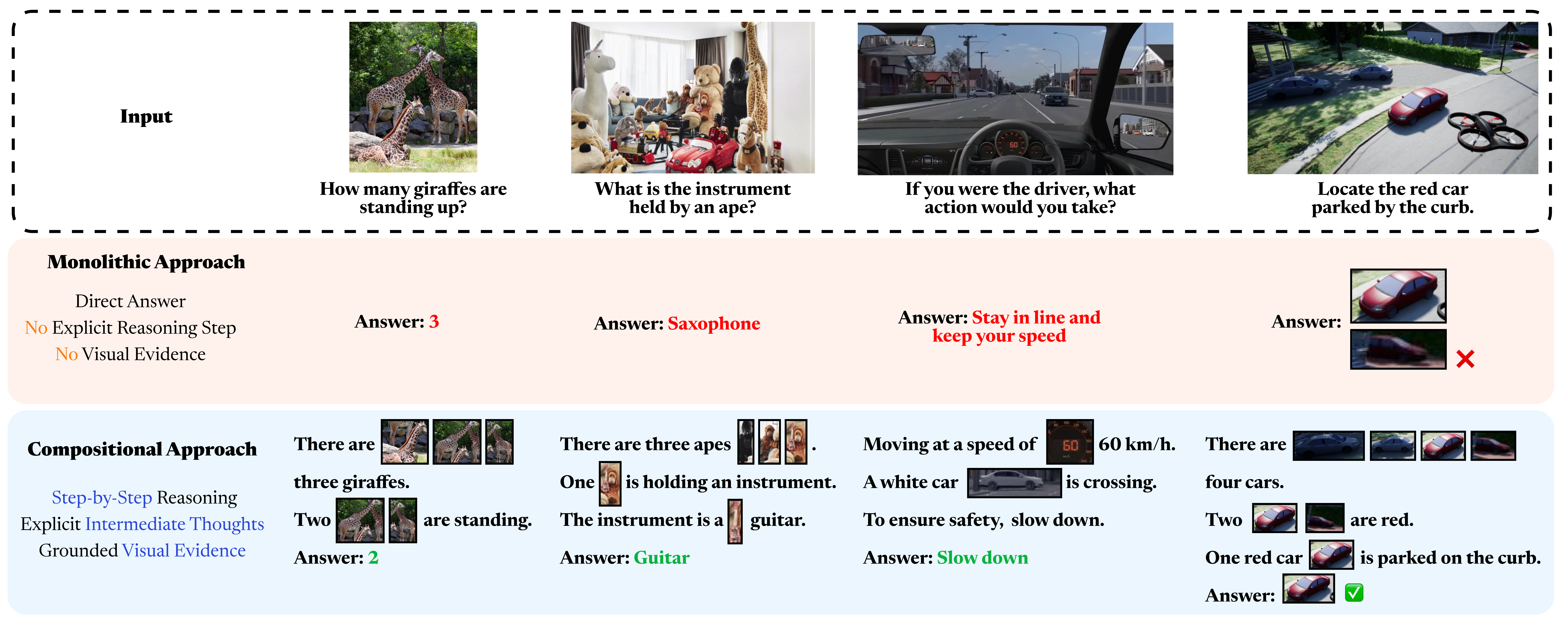} 
\end{center}
\vspace{-1.5em}
\caption{Illustration of visual reasoning tasks, highlighting the differences between monolithic and compositional approaches. Monolithic models map input to output directly, which often leads to hallucinations or incorrect outputs due to the lack of intermediate reasoning. In contrast, compositional methods explicitly break down the task into a sequence of interpretable reasoning steps. Each step is grounded in visual evidence, allowing the model to progressively infer the answer with greater transparency, accuracy, and robustness. 
}
\label{fig:teaser}
\end{figure*}

Humans possess a remarkable ability to interpret high-dimensional, uncompressed visual input and abstract its underlying structure, enabling the manipulation of underlying concepts as symbolic representations with notable efficiency~\citep{zerroug2022benchmark_cvr, zhang2024takeAstepback}. As a result, humans can effortlessly locate a target object in a cluttered room, determine whether a cup will fit inside a drawer, or predict the direction in which a stack of objects might fall. This cognitive capability, known as \textit{visual reasoning}, is widely regarded as a concentrated embodiment of human intelligence, forming the basis for concept formation, world understanding, and interaction with the environment~\citep{zhang2024takeAstepback, Frappier2018TheBO, johnson2017inferring, johnson2017clevr}. At its core, visual reasoning is inherently compositional: it requires recognizing objects and attributes, grounding them in visual evidence, modeling relations among them, and recombining these elements to support new inferences~\citep{johnson2017clevr, johnson2017inferring, lake2017building}.

In pursuit of human-level intelligence, a growing body of research has sought to replicate visual reasoning capabilities in machines~\citep{Zhang2019RAVENAD, Suchan2017VisualEB, jing2022maintaining_cvr, zerroug2022benchmark_cvr}. Recent advances in vision-language pretraining and large multimodal models have greatly expanded the scope of machine visual reasoning, enabling end-to-end systems to map visual and textual inputs directly to answers across natural images and embodied environments~\citep{liu2023llava, Qwenvl2023, li2023blip2, dai2023instructblip, liang2024surveyofmllm}. 
These systems have shown impressive results in general-purpose multimodal understanding~\citep{he2021interpretable_survey, bandyopadhyay2025thinking, chen2025towards_survey} and have been increasingly adopted across real-world applications~\citep{ma2025position, liu2023llava, fernandez2015uav, rogers2019vr_medical, zhan2020vr_medical, wang2023programmatically}.

Despite the promising performance of monolithic approaches in general multimodal understanding, significant challenges remain, particularly when viewed from the perspective of the inherently compositional and multi-step nature of human visual reasoning~\citep{thrush2022winoground, yuksekgonulBadofWord, Kafle2019ChallengesAP, liang2024surveyofmllm, wang2024reasoning_mllm_survey, man2025argus, yang2025thinkinginspace}. Concretely, we identify the following key limitations of monolithic approaches:
\begin{itemize}[leftmargin=3em]
    \item \textbf{Bias-driven shortcuts.} Instead of performing grounded inference, monolithic models often exploit spurious correlations and linguistic priors to produce plausible but incorrect answers~\citep{thrush2022winoground, yuksekgonulBadofWord}, undermining their generalization to challenging or novel scenarios.
    
    \item \textbf{Poor scaling with reasoning complexity.} Scaling up data and compute does not yield proportional gains when tasks demand multi-hop reasoning, spatial understanding, or precise grounding~\citep{ke2024hydra, stanictowards, wang2025multimodal_cot_reasoning_survey, sahin2024mmCR, deitke2025molmo}.
    
    \item \textbf{Weak compositional generalization.} Holistic models struggle to recombine objects, attributes, and relations in novel configurations, limiting their alignment with human-like visual reasoning~\citep{johnson2017clevr, lake2017building, johnson2017inferring, chen2021meta_cvr, zheng2021improving, hsu2023left}.
    Monolithic models, by contrast, treat the input holistically and lack mechanisms for explicit decomposition and relational reasoning.

\end{itemize}
These limitations highlight the need for modular, interpretable, and compositionally aligned approaches to visual reasoning.


Motivated by these cognitive insights and the limitations of monolithic models, recent research has increasingly explored \textbf{Compositional Visual Reasoning (CVR)}, a paradigm that decomposes visual tasks into grounded, interpretable reasoning steps guided by visual perception. 
Rather than relying on direct end-to-end prediction, CVR aims to make intermediate evidence and inference processes explicit, enabling models to reason over objects, attributes, relations, actions, and external knowledge in a structured manner~\citep{johnson2017inferring, ke2024hydra}. 
As illustrated in Figure~\ref{fig:teaser}, this survey focuses on methods that expose or construct such intermediate reasoning processes before answer generation.

Given its rapid progress since 2023, compositional visual reasoning has emerged as a central paradigm in the study of visual intelligence. However, its development trajectory, methodological foundations, and future directions remain insufficiently examined. Existing surveys in visual reasoning have primarily focused on conventional or monolithic approaches~\citep{kim2025visualsurvey, manmadhan2020visual_survey, kuang2025_natural_language_vqa_survey}, and thus fail to capture the emerging trend of compositional visual reasoning, which has quickly become a core research direction.
Meanwhile, surveys focused on multimodal large language models and reasoning tend to emphasize general-purpose reasoning~\citep{lin2025mm_reasoning_survey, wang2025multimodal_cot_reasoning_survey}, neurosymbolic frameworks~\citep{khan2025_neurosymbolic_VR_survey}, abstract pattern recognition~\citep{malkinski2025abstrct_vr_survey}, or agent-based architectures~\citep{ xie2024mllm_agent_survey, wang2025multimodal_cot_reasoning_survey}.
Although related surveys cover aspects such as planning, perception, and symbolic abstraction, they do not systematically examine CVR as a distinct paradigm. 
As shown in Table~\ref{tab:survey_table}, a dedicated review of this area is still lacking.

\arrayrulecolor{lightgray}
\begin{table*}[ht]
\centering
\footnotesize
\caption{Overview of survey papers on visual reasoning and multimodal AI since 2024. While informative, these surveys largely overlook the recent surge of CVR methods, highlighting the need for a dedicated review as pursued in this work.}
\label{tab:survey_table}
\begin{tabularx}{\textwidth}{
    >{\raggedright\arraybackslash}p{2.6cm}
    >{\raggedright\arraybackslash}p{3cm}
    X
}
\toprule
\textbf{Survey}  & \textbf{Topic} & \textbf{Contribution} \\

\midrule
\cite{ishmam2024imagesurvey}, \cite{kim2025visualsurvey}& Feature extraction and vision language pre-training for VQA & Critically analyzes feature extraction and vision language pre-training techniques of VQA, introduces a taxonomy, discusses dataset/method trends, and highlights emerging challenges and application opportunities. \\

\midrule
\cite{ma2024robustsurvey} & Robustness in VQA under distribution shift & Focuses on robustness in VQA across in- and out-of-distribution settings. Proposes typology for debiasing techniques and assesses model reliability and dataset quality.  \\

\midrule
\cite{wang2024reasoning_mllm_survey} & Reasoning capabilities in MLLMs & Categorizes multimodal large language models and their reasoning types (deductive, abductive, analogical). Summarizes evaluation protocols and trends in reasoning-focused applications. \\

\midrule
\cite{xie2024mllm_agent_survey}& LLM-driven multimodal agent & Systematically reviews LLM-driven agents, proposes taxonomies for perception and action, and introduces unified evaluation frameworks and application landscapes. \\

\midrule
\cite{lin2025mm_reasoning_survey} & Reasoning capabilities in MLLMs & Categorizes reasoning into language-centric and collaborative modes. Discusses technical challenges like semantic alignment and dynamic interaction. \\

\midrule
\cite{wang2025multimodal_cot_reasoning_survey} & Multimodal CoT reasoning & First survey dedicated to Multimodal CoT reasoning. Provides taxonomy, benchmarks, challenges, and resources to support research in multimodal AGI. \\

\midrule
\cite{khan2025_neurosymbolic_VR_survey} & Neurosymbolic reasoning with scene graphs and commonsense knowledge & First to survey neurosymbolic visual reasoning combining deep learning, scene graphs, and common sense knowledge. Categorizes methods by architecture and knowledge type, and outlines key challenges. \\

\midrule
\cite{li2025perception_reason_survey} & Large multimodal reasoning models & Provides a comprehensive survey of large multimodal reasoning models. Proposes a three-stage road-map (modular, multimodal CoT, System-2), introduces native large multimodal reasoning models, and reorganizes benchmarks and datasets for multimodal reasoning. \\

\midrule
\cite{kuang2025_natural_language_vqa_survey} & Natural language understanding and inference within VQA & Reviews VQA models focusing on perceptual and cognitive reasoning. Highlights recent progress in few-shot VQA and knowledge-guided inference in MLLMs. \\

\midrule
\cite{song2025bridge} & Modality alignment methods in MLLMs & Categorizes four alignment methods in MLLMs (converter, perceiver, tool, data-driven). Traces evolution and highlights key challenges in multimodal semantic bridging. \\

\midrule
\cite{qu2025tool_llm_survey} & Tool learning with LLM & Surveys tool learning from motivation to implementation. Defines four tool-use stages and discusses benchmarks, evaluation, and design challenges. \\

\midrule
\cite{malkinski2025abstrct_vr_survey} & Abstract visual reasoning & Reviews deep learning methods for solving Raven’s Progressive Matrices. Provides benchmarks,  comparisons, and insights.\\

\bottomrule
\end{tabularx}
\end{table*}
\arrayrulecolor{black}

To address this gap, we primarily focus on image-centered compositional visual reasoning, while discussing 3D, multi-view, embodied, and world-model-based methods when they directly extend or illuminate image-based CVR mechanisms. We focus on methods published between January 2023 and June 2026 that introduce explicit intermediate reasoning steps before answer generation, including LLM-guided inference, tool-integrated workflows, chain-of-thought-based visual reasoning, and agentic VLM architectures. Video reasoning and 3D reasoning are not treated as standalone survey targets, but are discussed when they directly inform the methodological development of image-based CVR. 
The main contributions of this survey are fourfold. 
\begin{enumerate}[leftmargin=2em]
    \item \textbf{Systematic motivation for CVR.}
    We clarify why CVR is needed for modern multimodal systems, focusing on grounding, compositional generalization, interpretability, robustness, and efficiency.

    \item \textbf{A stage-wise taxonomy of CVR methods.}
    We propose a five-stage taxonomy of large-model-enhanced CVR methods, spanning prompt-based reasoning, tool-augmented workflows, grounded chain-of-thought, and agentic VLMs.

    \item \textbf{Comprehensive review of benchmarks and evaluation.}
    We summarize representative datasets and evaluation settings used in CVR research, highlighting their task types, visual inputs, reasoning targets, and evaluation metrics.

    \item \textbf{Open challenges and future directions.}
    We synthesize key bottlenecks and outline future directions toward grounded, interpretable, robust, and compositionally generalizable visual reasoning.
\end{enumerate}


The remainder of this survey is organized as follows. 
Sec.~\ref{sec:background} introduces key definitions, including visual reasoning, monolithic visual reasoning, and compositional visual reasoning. 
Sec.~\ref{sec:reason} discusses why CVR is needed, covering cognitive alignment, semantic and relational understanding, robustness, interpretability, modular reuse, bias mitigation, and data efficiency. 
Sec.~\ref{sec:shift} presents our multi-stage taxonomy of CVR methods, from prompt-enhanced LLM-centric pipelines to tool-augmented systems, CoT-based VLMs, and unified agentic VLMs. 
Finally, we summarize key challenges, and discuss future directions in Sec.~\ref{sec:challenge}.

\section{Background}
\label{sec:background}

In this section, we provide an overview of the key concepts and terminology related to compositional visual reasoning, offering a clearer understanding of the fundamental aspects of this field.

\subsection{Visual Reasoning}
Visual reasoning is a cognitive process that enables the interpretation and analysis of relationships among entities within a visual scene to support decision-making and problem-solving~\citep{malkinski2025abstrct_vr_survey, wang2024reasoning_mllm_survey, he2021interpretable_survey, cocchi2025augmenting}. Rather than being limited to low-level perception, visual reasoning involves the integration of high-level information from multiple modalities (\eg, vision and language) to facilitate deeper semantic understanding~\citep{ishmam2024imagesurvey, kuang2025_natural_language_vqa_survey}. 
This multimodal reasoning underlies tasks such as visual question answering, entailment, and grounding~\citep{zhang2020multimodal, hong2021transformation}, where accurate comprehension and logical analysis of visual content are critical.

In this survey, we focus on visual reasoning tasks that involve interpreting visual inputs (\ie, images) paired with textual queries or prompts to produce structured outputs. Formally, we denote the visual input as $v$, the query or prompt as $q$, and the answer as $y$. A visual reasoning system is a function $\mathcal{M} : (v, q) \mapsto y$, which varies depending on the reasoning approach, monolithic or compositional. The form of $y$ also varies depending on the task and may include natural language responses, discrete answer selections (\eg, multiple choice), bounding boxes, or segmentation masks. Representative tasks include, but are not limited to, general visual question answering (\eg, ``What color is the car?"), visual grounding (\eg, ``Find the chair next to the window"), relational scene graph reasoning (``Is the cat under the table?").
 
At its core, visual reasoning comprises two essential components: visual perception, which involves extracting visual information such as objects, attributes, and relations~\citep{he2016resnext, he2025Two_stage}, and logical reasoning, which applies structured reasoning to derive conclusions based on the perceived information~\citep{he2023hierarchical, redmon2016yolo}. While intuitive for humans, this capability remains a major challenge for artificial intelligence systems due to the requirement for high-level abstraction and generalization.


\subsection{Monolithic Visual Reasoning}
Monolithic visual reasoning methods are a class of neural network models $\mathcal{M}$ that directly map visual input $v$ and textual query $q$ to an output answer $y$, where $\mathcal{M}$ commonly directly encodes both modalities into a joint representation and predicts the answer without exposing intermediate steps. These architectures do not explicitly represent reasoning steps, performing reasoning implicitly within end-to-end pipelines~\citep{zerroug2022benchmark_cvr, chen2021meta_cvr}. 
These models typically extract visual features from the entire image $v$ and combine them with language embeddings using co-attention or joint encoding to perform implicit multimodal reasoning~\citep{yu2017multi, kim2018bilinear, liu2023llava, li2025perception_reason_survey}.

Early models such as MFB~\citep{yu2017multi}, and MCAN~\citep{yu2019deep} used convolutional features and cross-modal attention for direct answer prediction. Recent extensions include dual-encoder contrastive models (\eg, ViLBERT~\citep{lu2019vilbert}, CLIP~\citep{radford2021clip}, BLIP2~\citep{li2023blip2}), single-Transformer backbones (\eg, UNITER~\citep{chen2020uniter}, Flamingo~\citep{alayrac2022flamingo}), and vision-encoder-to-LLM pipelines (\eg, LLaVA~\citep{liu2023llava}, MiniGPT-4~\citep{zhu2024minigpt}, InstructBLIP~\citep{dai2023instructblip}, Qwen-VL~\citep{Qwenvl2023}).

\subsection{Compositional Visual Reasoning}

Compositional reasoning approaches map an input pair $(v, q)$ to an answer $y$ via an intermediate structured representation, for example, a sequence of $n$ steps $S = \{s_1, s_2, \dots, s_n\}$, where the decomposition is either inferred dynamically or predefined based on the query structure. Here, $S$ is used as an abstract representation of the reasoning trajectory rather than a fixed implementation form. Each step $s_i$ may be instantiated as a symbolic operation, an LLM-generated instruction, a multimodal model invocation, or a call to an external visual tool, depending on the framework. When grounded to the visual input $v$, a step may involve finer-grained operations such as object identification ($o_i$), attribute inference ($a_i$), and relation resolution ($r(o_i, o_j)$). These grounding processes may encompass visual perception capabilities such as object detection, attribute recognition, and depth estimation accordingly. The steps $S$ can then be executed sequentially or composed hierarchically into a symbolic program, ultimately enabling the model to produce the answer $y$.


Compositional visual reasoning is a specialized form of visual reasoning that emphasizes such structured decomposition and recombination of visual elements to solve complex tasks~\citep{ma2023crepe}. It is grounded in the principle of compositionality, which posits that ``the meaning of the whole is a function of the meanings of its parts''~\citep{cresswell2016logics_and_language}. Rather than treating a scene as a monolithic input, compositional reasoning explicitly identifies and manipulates objects, attributes, and their interrelations to construct flexible reasoning pathways~\citep{chen2021meta_cvr, zerroug2022benchmark_cvr, jing2022maintaining_cvr}. 

A hallmark of compositional visual reasoning is its capacity for systematic generalization: the ability to infer novel combinations of familiar elements without direct exposure to those configurations during training~\citep{chen2021meta_cvr, johnson2017inferring}. Concretely, this generalization refers to executing reasoning over the combinatorial product space of primitive visual elements, such as objects $o$, attributes $a$, and relations $r(o_i, o_j)$, as well as adapting to different orderings of steps $S$. This mirrors human-like cognitive flexibility and is central to building AI systems that are interpretable, modular, and capable of robust generalization across diverse multimodal tasks~\citep{ke2024hydra, stanictowards}.

\section{Why Compositional Visual Reasoning?}
\label{sec:reason}

In this section, we outline the core motivations for adopting compositional visual reasoning, emphasizing its advantages over traditional monolithic models. Specifically, we examine six key perspectives: cognitive alignment with human reasoning; semantic and relational understanding; generalization and robustness; transparency, interpretability, and modular reuse; reducing language biases and hallucinations; and lower data requirements and improved efficiency.

\subsection{Cognitive Alignment with Human Reasoning}
Compositional visual reasoning (CVR) is inspired by the systematic and modular way humans approach visual problems~\citep{ma2023crepe, lake2017building, zellers2019recognition_VCR, gupta2023visprog, hudson2019gqa, lu2023chameleon}. Humans naturally break down complex scenes into interpretable components (\eg, objects, attributes, and relationships) and apply sequential reasoning and contextual adaptation~\citep{zerroug2022benchmark_cvr, jiang2022bongard_hoi, ke2024hydra, ke2025dwim}. This process reflects our ability to manipulate abstract symbols and relational concepts extracted from visual inputs.

Moreover, the human capacity for compositional reasoning, rapid concept formation, and relational generalization is particularly remarkable~\citep{lu2023chameleon, chen2021meta_cvr, jing2022maintaining_cvr,Zheng_2024_CVPR}. These abilities, often associated with fluid intelligence and non-verbal reasoning, enable flexible understanding across novel contexts and support exceptional sample efficiency when learning new visual concepts. Therefore, a key desideratum for CVR models is to mirror core cognitive strategies, such as task decomposition, focused attention, iterative refinement, and the use of external tools to isolate critical visual details~\citep{zhang2025critic, liao2025canvlmcorr}.

\subsection{Semantic and Relational Understanding}
Traditional monolithic models often rely on superficial correlations and pattern matching, which limits their ability to capture deep semantic structures or precise object-centric relationships. In contrast, compositional visual reasoning enables explicit modeling of such relationships through structured representations like scene graphs or deductive reasoning within the language space~\citep{wang2024reasoning_mllm_survey, khan2025_neurosymbolic_VR_survey, hou2020VRlearning, stone2025semanticgui, zengsocratic, park2025svg}. This allows systems to reason about spatial configurations, semantic relations, and attribute interactions in a principled and interpretable manner~\citep{ma2024robustsurvey, khan2025_neurosymbolic_VR_survey}. Such capabilities are essential for abstract scene understanding tasks, including counting, identifying intersections, and comparing object properties~\citep{lin2025mm_reasoning_survey}. Moreover, compositional reasoning facilitates multimodal understanding by bridging vision and language, enabling accurate interpretation of complex visual elements in diagrams and charts~\citep{wang2025multimodal_cot_reasoning_survey}.

\subsection{Generalization and Robustness}
Monolithic visual reasoning models are often brittle and prone to bias, with poor performance on out-of-distribution or few-shot tasks due to shortcut learning~\citep{johnson2017clevr, chen2024p2g, hsu2023disco}. Compositional approaches, by contrast, promote systematic generalization—the ability to solve novel tasks by recombining known elements, such as objects, attributes, or relations~\citep{chen2023divide, johnson2017inferring}. For example, when structured models learn to ground visual concepts in a scene (\eg, ``monitor'' and ``desk'') and resolve relations (\eg, ``on top of''), they can generalize to identify any object-relation pair (\eg, monitor on desk), without having seen that specific combination during training. This ability reduces dependence on dataset-specific biases and enhances robustness across domains. Benchmarks like CLEVR~\citep{johnson2017clevr} and VCR~\citep{zellers2019recognition_VCR} have demonstrated the necessity of compositional reasoning to prevent models from exploiting statistical priors and encourage faithful reasoning~\citep{zerroug2022benchmark_cvr, jiang2022bongard_hoi}.

\subsection{Transparency, Interpretability, and Modular Reuse}

Compositional visual reasoning enables the generation of intermediate outputs (\eg, bounding boxes, scene graphs, textual rationales) that expose the model’s decision path, improving interpretability and allowing targeted debugging~\citep{gupta2023visprog, li-etal-2025-vocot}. This ability is essential for building trustworthy and reliable systems, especially in domains where high precision is required. In contrast to black-box architectures, compositional systems are modular: components can be reused across tasks, reducing the need to retrain models from scratch and enabling plug-and-play integration of vision experts~\citep{khan2024visrep, cheng2024vireo, su2025thinkingsurvey}. As individual components improve, so does the full compositional framework. This flexibility also leads to efficiency gains in inference and model adaptation.

\subsection{Reducing Language Biases and Hallucinations}
While generalization focuses on performance across diverse input distributions, another critical challenge for VLMs lies in the generation of inaccurate or unsupported content. 
Monolithic methods, especially those trained to directly map visual inputs to answers or rationales, are highly susceptible to hallucinations. These models might generate responses that appear plausible but are actually incorrect or ungrounded, particularly when they depend too heavily on linguistic priors~\citep{zhou2024imageofthought, yang2023dawn, Wu_2026_CVPR_mmgcot}. For instance, when asked about the color of a green banana, a monolithic model might incorrectly answer ``yellow." This issue stems from shortcut learning, where models exploit statistical biases in the training data rather than grounding their reasoning in visual evidence~\citep{johnson2017clevr, chen2024p2g, li2024naturalbench, fu2024blink}.
In contrast, compositional reasoning mitigates this by enforcing explicit grounding: intermediate reasoning steps, such as object detection, relation extraction, or visual token selection, compel the system to anchor its inferences in actual visual content. This not only improves factual consistency but also reduces reliance on spurious correlations and enhances robustness in multimodal reasoning tasks~\citep{ishmam2024imagesurvey, li-etal-2025-vocot, wu2024conceptmix}.
Furthermore, compositional methods can incorporate external knowledge or common sense when needed, but do so in a controlled and interpretable manner, avoiding the conflation of visual and linguistic content.



\subsection{Lower Data Requirements and Improved Efficiency}
Traditional monolithic models often require massive datasets and computational resources to achieve acceptable performance, particularly as training must account for the combinatorial explosion of possible visual-linguistic combinations~\citep{ke2024hydra, stanictowards}. This approach is not only costly but also inefficient, especially in dynamic or specialized domains.
In contrast, compositional visual reasoning promotes modularity and reusability: once basic visual skills (\eg, object detection, OCR, segmentation) are learned, they can be reused across tasks without retraining~\citep{kimvisual_reasoning}. This modular structure reduces the demand for extensive annotated datasets and enables few-shot or zero-shot generalization to new tasks by recombining known components~\citep{jiang2022bongard_hoi, zerroug2022benchmark_cvr, jin2025predicate}.
Additionally, techniques such as visual program distillation or intrinsic tool orchestration (\eg, PaLI~\citep{hu2024pali}, Griffon-R~\citep{zhan2025griffonr}) allow compositional systems to approximate multi-step reasoning with reduced runtime and compute overhead in deployment~\citep{hu2024pali, zhan2025griffonr}. 
Together, these advantages position compositional visual reasoning as a promising paradigm for building more capable, trustworthy, and efficient multimodal systems that align more closely with human reasoning.
\section{Key Stages of Compositional VR Paradigms}
\label{sec:shift}

\begin{figure*}[t]
\begin{center}
  \includegraphics[width=0.95\linewidth]{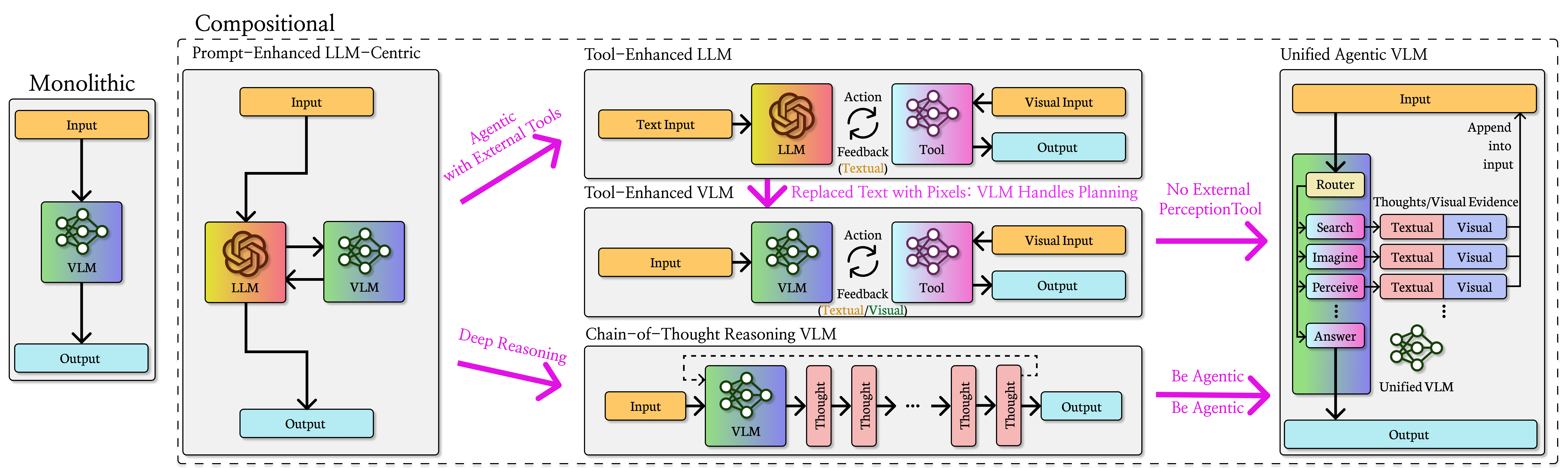} 
\end{center}
\vspace{-1.5em}
\caption{Key shift from monolithic reasoning to compositional reasoning. }
\label{fig:relation}
\vspace{-1em}
\end{figure*}

\begin{table*}[t]
\centering
\small
\setlength{\tabcolsep}{7.2pt}
\renewcommand{\arraystretch}{1.16}
\caption{
Capability-level comparison of the five compositional visual reasoning paradigms.
}
\label{tab:stage_summary_compact}

\resizebox{0.95\textwidth}{!}{%
\begin{tabular}{
>{\raggedright\arraybackslash}p{0.10\textwidth}
>{\centering\arraybackslash}p{0.10\textwidth}
>{\centering\arraybackslash}p{0.08\textwidth}
>{\centering\arraybackslash}p{0.08\textwidth}
>{\centering\arraybackslash}p{0.095\textwidth}
>{\centering\arraybackslash}p{0.08\textwidth}
>{\centering\arraybackslash}p{0.08\textwidth}
>{\raggedright\arraybackslash}p{0.20\textwidth}
}
\toprule
\textbf{Paradigm}
& \textbf{Reasoner}
& \textbf{Direct Vision}
& \textbf{External Tools}
& \textbf{Visual Feedback}
& \textbf{Internal CoT}
& \textbf{Agentic Loop}
& \textbf{Main Bottleneck} \\
\midrule

\rowcolor{rowblue}
\textbf{Stage I}
& LLM
& \xmark
& \xmark
& \xmark
& \pmark
& \xmark
& Weak grounding \\

\textbf{Stage II}
& LLM
& \xmark
& \cmark
& \pmark
& \pmark
& \cmark
& Tool coordination \\

\rowcolor{rowblue}
\textbf{Stage III}
& VLM
& \cmark
& \cmark
& \cmark
& \pmark
& \cmark
& Tool dependency \\

\textbf{Stage IV}
& VLM
& \cmark
& \xmark
& \xmark
& \cmark
& \xmark
& Single-pass reasoning \\

\rowcolor{rowblue}
\textbf{Stage V}
& VLM
& \cmark
& \xmark
& \cmark
& \cmark
& \cmark
& Cost; supervision \\

\bottomrule
\end{tabular}%
}

\vspace{2pt}
\footnotesize{
\cmark: explicit support; \pmark: partial or optional support; \xmark: generally absent.
}
\end{table*}

\begin{figure*}[t]
\begin{center}
  \includegraphics[width=0.97\linewidth]{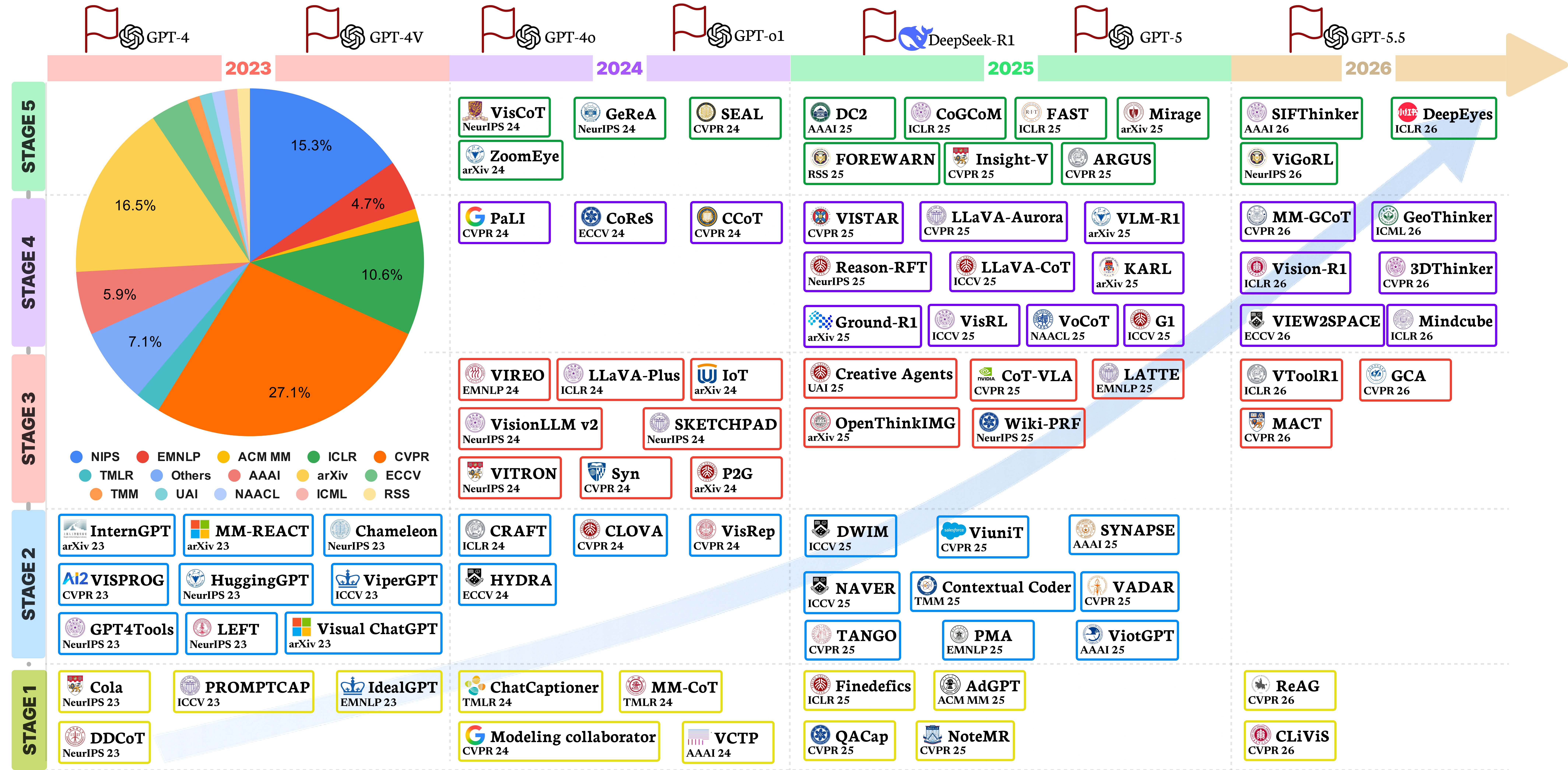} 
\end{center}
\vspace{-1em}
\caption{Roadmap of compositional visual reasoning models. }
\label{fig:roadmap}
\vspace{-1em}
\end{figure*}

In this section, we trace the evolution of compositional visual reasoning models in recent years, highlighting how each stage has progressively enhanced their ability to decompose complex queries, reason over structured visual elements, and generalize across novel compositions. These transitions, illustrated in Figure~\ref{fig:relation}, capture the key methodological shifts that have shaped modern compositional visual reasoning systems.

Early monolithic large vision-language models like ~\citep{liu2023llava, alayrac2022flamingo, wang2024qwen2, xiao2024florence, liu2024llavanext, team2024gemini, zhu2024minigpt, lu2024deepseekvl, li2023blip2, dai2023instructblip, achiam2023gpt4v, ye2023mplug} adopt end-to-end architectures that directly map visual and textual inputs to answers. These models aim to learn human-like visual reasoning capabilities directly from data and perform well on perceptual and shallow reasoning tasks. However, they struggle with more complex visual reasoning challenges, such as spatial understanding and fine-grained grounding~\citep{chen2024spatialvlm, liu2023vspatialreasoning, qi2024cogcom}. These limitations have driven the development of compositional visual reasoning approaches, which decompose the reasoning process into structured, interpretable steps. Figure~\ref{fig:relation} and Table~\ref{tab:stage_summary_compact} illustrate the key architectural transition from monolithic to compositional paradigms since $2023$, and Figure~\ref{fig:roadmap} highlights representative models at each stage of this evolution.
The recent surge of interest in compositional visual reasoning is largely rooted in the growing recognition of LLMs' powerful reasoning capabilities in domains such as natural language, mathematics, and code~\citep{austin2021program, kojima2022large, cobbe2021gsm8}. These advances have inspired researchers to explore whether such symbolic and structured reasoning abilities can be extended to vision-language settings.

Recognizing monolithic approaches' limitations, researchers began exploring whether the structured reasoning strengths of LLMs could be leveraged for visual domains. Given LLMs’ ability to decompose and solve complex problems in language, math, and code, a natural question arose: 
\begin{tcolorbox}
[boxsep=0mm,left=1.2mm,right=1.2mm,colframe=black!55,colback=black!5]
{\textbf{Question I:} Can complex tasks be decomposed into a sequence of simpler sub-questions, each solvable by a VLM, followed by answer synthesis via an LLM?}
\end{tcolorbox}
\textbf{Stage I: Prompt-Enhanced Language-Centric Methods} provides an effective solution to \textbf{Question I}.
These approaches leverage LLMs' strong language-based reasoning ability to break down complex visual questions into a series of simpler sub-questions, each of which can be independently answered by a VLM. After obtaining the sub-answers, the LLM performs final reasoning and synthesis entirely in the language space to produce the overall answer.
This idea has given rise to compositional visual reasoning as a prominent research direction. Rather than treating reasoning as a single-step mapping from input to output, compositional visual reasoning paradigms frame it as a multi-stage process involving perception, task decomposition, intermediate reasoning, and final synthesis, as illustrated in Figure~\ref{fig:relation}.

Although Stage I prompt-enhanced language-centric methods have achieved some success, they remain limited in visual understanding due to weak grounding in the underlying visual content and their relatively fixed prompting logic. This naturally raises an important question: 
\begin{tcolorbox}
[boxsep=0mm,left=1.2mm,right=1.2mm,colframe=black!55,colback=black!5]
{\textbf{Question II:} Can complex visual reasoning tasks be addressed more effectively by enabling models to actively invoke and coordinate multiple external tools through a central decision-making module?}
\end{tcolorbox}
\textbf{Stage II: Tool-Enhanced LLMs} have emerged as a new paradigm to solve \textbf{Question II}.
In this paradigm, the LLM issues grounding actions to external tools, and an external environment executes these actions to extract relevant information from the visual input. This interaction enables the system to progressively gather visual evidence and address complex visual reasoning tasks more effectively.

These methods offer greater reasoning flexibility, but often struggle with effective tool coordination, frequently resulting in noisy or suboptimal workflows. In most cases, the tool action planner, usually an LLM, has access to visual information only in the form of text descriptions produced by upstream vision models. If these textual descriptions are incomplete or inaccurate, the reasoning quality degrades and cannot be fully recovered downstream. This raises a natural question: 
\begin{tcolorbox}
[boxsep=0mm,left=1.2mm,right=1.2mm,colframe=black!55,colback=black!5]
{\textbf{Question III:} Can VLMs, which directly perceive visual inputs rather than relying on textual proxies, provide improved performance and greater structural advantages in tool-enhanced frameworks for visual reasoning?}
\end{tcolorbox}
\textbf{Stage III: Tool-Enhanced VLMs} have emerged consequently as an answer to \textbf{Question III}, allowing the planner to access visual input directly rather than relying solely on textual descriptions. This direct perception reduces semantic loss, enables more accurate and informed action planning, and opens new possibilities for designing both the format of grounding actions and the structure of visual feedback. In some cases, this enables models to simulate visual imagination and perform visual verification by generating or manipulating visual content via tools during the reasoning process.

While prompt-enhanced and tool-enhanced methods rely on explicit problem decomposition and inter-module interaction to support compositional reasoning, this motivates the following question:
\begin{tcolorbox}
[boxsep=0mm,left=1.2mm,right=1.2mm,colframe=black!55,colback=black!5]
{\textbf{Question IV:} Can VLMs directly perform perception and generate reasoning steps to derive an answer?}
\end{tcolorbox}
\textbf{Stage IV: Chain-of-Thought VLMs} are a promising solution to the challenges posed in \textbf{Question IV}, adopting unified architectures that tightly integrate perception and inference. These models perform multi-step reasoning without relying on external tools, and explicitly reveal intermediate thought processes and perception information before the final answer.

While chain-of-thought (CoT) VLMs integrate perception and reasoning, they remain constrained by relatively rigid pipelines, single-pass forward architectures, and reliance on textual representations of thought. Rather than directly interacting with visual content (\eg, observing specific image regions), these models typically convert visual information into text and reason over it, limiting their ability to dynamically perceive and reflect. Based on these trends, a critical research question arises:
\begin{tcolorbox}
[boxsep=0mm,left=1.2mm,right=1.2mm,colframe=black!55,colback=black!5]
{\textbf{Question V:} Can VLMs behave more like humans by autonomously exploring, inspecting, and manipulating visual content while reasoning iteratively and adaptively, all without relying on external tools?}
\end{tcolorbox}
This question lies at the heart of \textbf{Stage V: Unified Agentic VLMs}, which represent the latest evolution in the field. These models incorporate higher-order cognitive mechanisms such as planning, memory, visual operation, imagination, textual feedback and visual evidence. Such capabilities enable iterative perception, step-by-step reasoning, and adaptive decision-making for complex visual tasks.

\subsection{Stage I: Prompt-Enhanced Language-Centric Methods}
\label{sec:promptllm}

\begin{figure*}[t]
\begin{center}
  \includegraphics[width=0.95\linewidth]{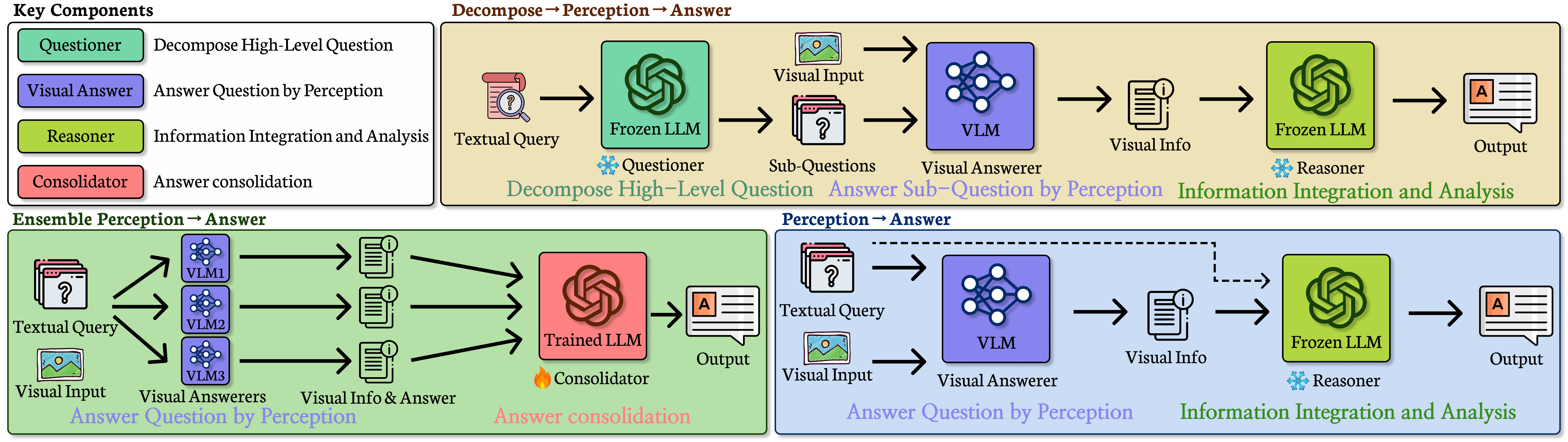} 
\end{center}
\caption{Overview of prompt-enhanced language-centric (Stage I) pipeline.}
\label{fig:stage1}
\end{figure*}

\begin{table*}[t]
\centering
\small
\setlength{\tabcolsep}{6.5pt}
\renewcommand{\arraystretch}{1.14}
\caption{
Representative prompt-enhanced language-centric methods in Stage I.
}
\label{tab:stage1_methods}

\resizebox{0.98\textwidth}{!}{%
\begin{tabular}{l l c l l l}
\toprule
\textbf{Model}
& \textbf{Venue}
& \textbf{Task Struct.}
& \textbf{Visual Grounding}
& \textbf{Synthesis}
& \textbf{Training} \\
\midrule

\rowcolor{rowblue}
DDCoT~\cite{zheng2023ddcot}
& NeurIPS 2023
& \cmark
& VLM sub-answers
& LLM
& Training-free \\

CoLa~\cite{chen2023cola}
& NeurIPS 2023
& \pmark
& Multi-VLM outputs
& LLM 
& Fine-tuned \\

\rowcolor{rowblue}
IdealGPT~\cite{you2023idealgpt}
& EMNLP 2023
& \cmark
& VLM sub-answers
& LLM
& Training-free \\

PromptCap~\cite{hu2023promptcap}
& ICCV 2023
& \xmark
& Question-aware captions
& LLM
& Fine-tuned \\

\rowcolor{rowblue}
ChatCaptioner~\cite{zhuchatgptask}
& TMLR 2024
& \cmark
& BLIP-2 QA
& LLM
& Training-free \\

Modeling collaborator~\cite{toubal2024modelingcollaborator}
& CVPR 2024
& \cmark
& VLM feedback
& LLM 
& Training-free \\

\rowcolor{rowblue}
AdGPT~\cite{AdGPT_huang}
& ACM MM 2024
& \pmark
& Dialogue-driven captions
& LLM
& Training-free \\

MM-CoT~\cite{zhang2024mmcot}
& TMLR 2024
& \cmark
& Visual rationale
& LLM
& Fine-tuned \\

\rowcolor{rowblue}
Finedefics~\cite{heFinedefics}
& ICLR 2025
& \pmark
& Fine-grained perception
& LLM
& Fine-tuned \\

VCTP~\cite{chen2024vctp}
& AAAI 2024
& \cmark
& See-think-confirm
& LLM
& Fine-tuned \\

\rowcolor{rowblue}
QACap~\cite{Yang_2025_CVPR_QACap}
& CVPR 2025
& \pmark
& Linguistic observation
& LLM
& Fine-tuned \\

NoteMR~\cite{Fang_2025_CVPR_NoteMR}
& CVPR 2025
& \pmark
& Visual notes
& VLM
& Fine-tuned \\

\rowcolor{rowblue}
ReAG~\cite{Compagnoni_2026_CVPR_REAG}
& CVPR 2026
& \pmark
& Retrieved evidence
& LLM 
& Fine-tuned  \\

CLiViS~\cite{Li_2026_CVPR_clivis}
& CVPR 2026
& \cmark
& VLM Cognitive map
& LLM
& Training-free \\

\bottomrule
\end{tabular}%
}

\vspace{2pt}
\footnotesize{
Task Struct. denotes explicit question decomposition, stage-wise prompting, or structured evidence organization.
\cmark: explicit support; \pmark: partial or implicit support; \xmark: generally absent.
}
\end{table*}

We introduce prompt-enhanced language-centric approaches in this subsection, with an overview illustrated in Figure~\ref{fig:stage1} and Table~\ref{tab:stage1_methods}. These methods capitalize on the reasoning capabilities of LLMs, which typically serve two complementary roles. First, as a task decomposition module, the LLM decomposes a complex visual question $q$ into a sequence of simpler sub-questions that can be independently addressed by VLMs given visual input $v$. Second, as a consolidator or reasoner, it synthesizes a final answer based on the perceptual outputs or intermediate answers provided by the VLMs. 


In this paradigm, the LLM performs the core reasoning process entirely within the language space, using information obtained from the VLM and textual query. The VLM is tasked with visual perception, either by directly answering decomposed sub-questions or by extracting relevant visual features from the input image to text descriptions. This modular design promotes interpretability, adaptability, and prompt-based control, and often functions effectively without requiring fine-tuning or task-specific supervision.

\subsubsection{Task Decomposition Followed by Visual Perception}
\label{sec:type1subtype1}
The typical architecture follows a structured pipeline in which a high-level visual question is first decomposed into a sequence of low-level sub-questions. Each sub-question is then independently addressed, and the final answer is synthesized based on the intermediate responses.
Early methods like DDCoT~\citep{zheng2023ddcot}, ChatCaptioner~\citep{zhuchatgptask}, IdealGPT~\citep{you2023idealgpt} and Modeling Collaborator~\citep{toubal2024modelingcollaborator} adopt a straightforward strategy in which a frozen LLM decomposes a complex visual question into sub-questions or instructions, which are then processed by pretrained VLMs such as BLIP-2~\citep{li2023blip2}. These approaches are lightweight and task-agnostic, enabling rapid prototyping and broad applicability. However, their reliance on fixed prompts and lack of coordination training often results in suboptimal reasoning paths, redundant queries, and brittle performance.

\subsubsection{Perceptual Grounding Prior to Reasoning}
To address the limitations mentioned in Section \ref{sec:type1subtype1}, more recent work introduces varying degrees of structured interaction. CoLa~\citep{chen2023cola} improves coordination by training the LLM to aggregate and reason over outputs from multiple VLMs, yielding better coherence and adaptability. PromptCap~\citep{hu2023promptcap}, MM-CoT~\citep{zhang2024mmcot}, AdGPT~\citep{AdGPT_huang}, and Finedefics~\citep{heFinedefics} shift the focus toward perception-first reasoning, where visual content is first converted into question-aware captions, dialogue-driven descriptions, visual rationales, or fine-grained evidence before LLM-based answer synthesis.

VCTP~\citep{chen2024vctp} strengthens this line by training the LLM to operate in a structured see-think-confirm pipeline. It adaptively grounds visual concepts, reasons over intermediate representations, and verifies outputs in a closed loop, improving transparency, consistency, and efficiency. CLiViS~\citep{Li_2026_CVPR_clivis} extends perception-first reasoning to embodied visual reasoning, where an LLM planner coordinates VLM-based perception to build a cognitive map and evidence memory for final answer synthesis.

Recent KB-VQA methods further instantiate this perception-first paradigm by separating visual observation, knowledge retrieval, and language-based reasoning. QACap~\citep{Yang_2025_CVPR_QACap}, NoteMR~\citep{Fang_2025_CVPR_NoteMR}, and ReAG~\citep{Compagnoni_2026_CVPR_REAG} respectively leverage question-aware observations, visual/knowledge notes, and reasoning-augmented retrieval to provide more reliable evidence for downstream LLM or MLLM reasoning. This separation enables compatibility with black-box LLMs and works well for knowledge-oriented VQA, but reasoning quality depends heavily on caption fidelity and lacks adaptive feedback.

\subsubsection{Synthesis and Outlook}
Overall, prompt-enhanced language-centric methods represent a clear progression from simple prompting strategies toward more coordinated and structured reasoning. However, these approaches remain limited in their visual grounding, as they rely solely on VLMs for perception without deeper integration of visual context. Additionally, their relatively fixed pipeline architecture constrains flexibility, making it difficult to adapt to more complex or dynamic visual reasoning tasks.

\subsection{Stage II: Tool-Enhanced Large Language Models}
\label{sec:toolllm}

In contrast to prompt-enhanced approaches that rely solely on textual reasoning, tool-enhanced LLMs equip language models with the ability to invoke external modules for perception, analysis, and computation. This paradigm typically involves two key components: \textbf{generating actions from the current state}, and \textbf{transitioning between states by executing those actions}. 

The LLM acts as a central planner or action generator, responsible for decomposing tasks, selecting appropriate tools, and synthesizing final answers~\citep{you2023idealgpt, ke2024hydra, zhong2025viotgpt}. Specifically, it maps the current state to grounding actions by producing structured tool calls, such as code snippets or formatted prompts. The environment then interprets and executes these actions, triggering state transitions on the visual input using corresponding tools. 
These tools may include vision models (\eg, detectors, captioners), knowledge bases, image editors, or code interpreters. Therefore, LLM's role is to understand tool descriptions, generate calls to those tools, and interpret returned results in a multi-turn or sequential manner.

\subsubsection{Single-Turn Framework}
\begin{figure*}[t]
\begin{center}
  \includegraphics[width=0.95\linewidth]{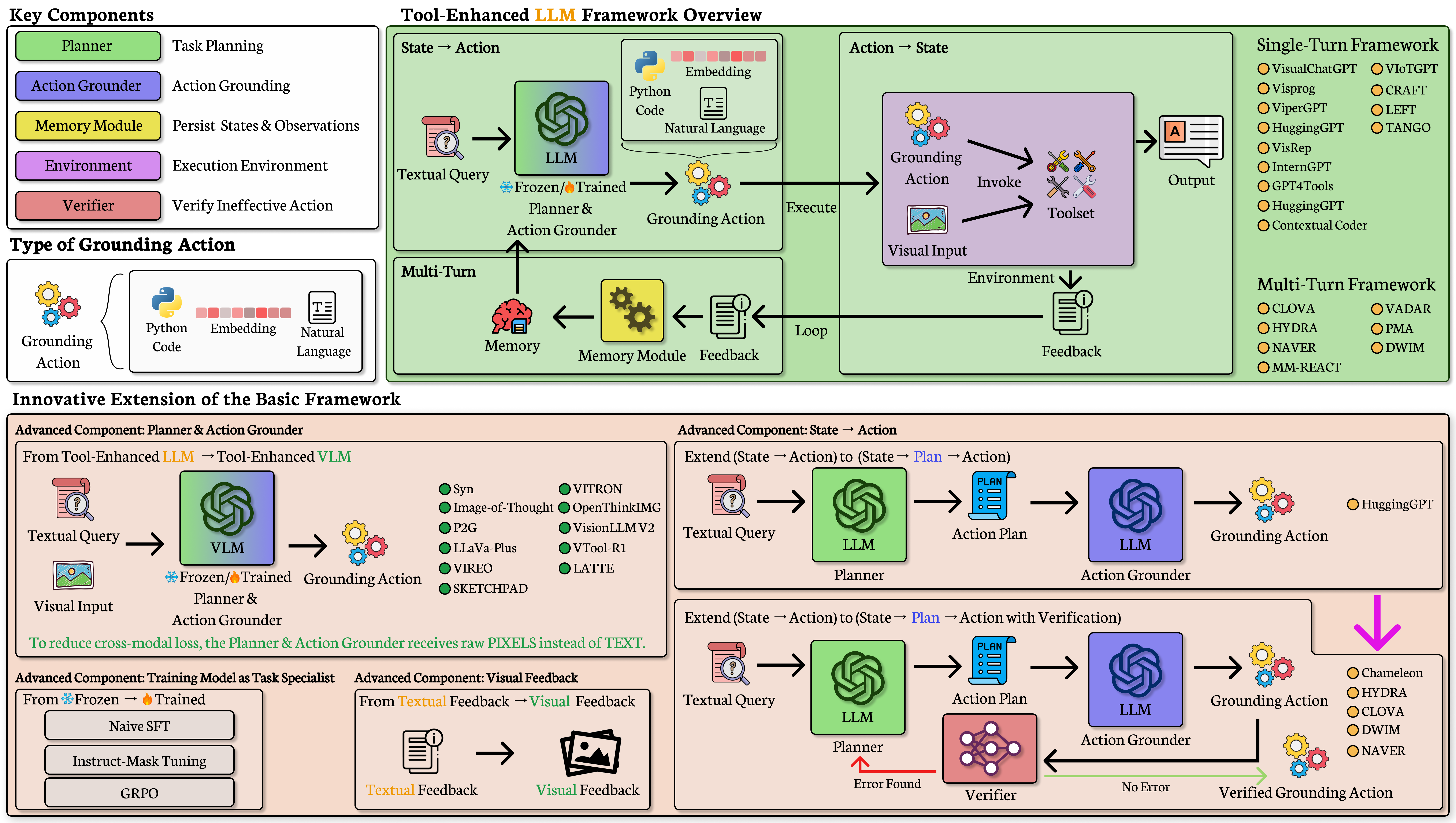} 
\end{center}
\vspace{-1em}
\caption{Tool-enhanced LLMs (Stage II) and VLMs (Stage III) pipeline.}
\label{fig:stage2}
\end{figure*}

Representative systems such as GPT4Tools~\citep{yang2023gpt4tools}, Visual ChatGPT~\citep{wu2023visualchatgpt}, Chameleon~\citep{lu2023chameleon}, VisProg~\citep{gupta2023visprog}, ViperGPT~\citep{suris2023vipergpt}, HuggingGPT~\citep{shen2023hugginggpt}, CRAFT~\citep{yuan2024craft}, Contextual Coder~\citep{shen2025contextualcoder} and InternGPT~\citep{liu2023interngpt} follow this structure, as illustrated in the tool-enhanced LLM framework overview in Figure~\ref{fig:stage2} and Table~\ref{tab:stage2_stage3_methods}. They assume access to a library of vision-language tools and use a frozen LLM to plan tool usage via prompt engineering, code generation, or natural language APIs. This setup enables training-free or zero-shot compositional reasoning across modalities, and provides a flexible interface to integrate new tools. However, these frozen LLMs often lack true tool awareness. They struggle to reason about tool capabilities, generate incorrect tool calls, and are unable to adapt when tool outputs are noisy or invalid~\citep{hsu2023visualscratchpad}.

\subsubsection{Adaptive Tool Use via Training, Feedback, and Verification}
While early tool-enhanced LLMs mainly rely on one-pass tool invocation, later methods shift toward adaptive tool use by introducing iterative feedback, tool-aware training, and structured executable reasoning.

One line of work closes the loop between tool execution and LLM reasoning. MM-REACT~\citep{yang2023mmreact} serializes vision-expert outputs as textual observations and feeds them back to ChatGPT, enabling iterative action-observation reasoning without updating the LLM itself. Building on this feedback-driven view, CLOVA~\citep{gao2024clova} and HYDRA~\citep{ke2024hydra} further turn tool use into an adaptive agentic process: CLOVA couples tool-augmented inference with reflection and continual tool updates, whereas HYDRA combines an LLM planner, a reinforcement-learning controller, and a reasoning engine to dynamically refine tool usage through interaction.

Another line improves tool awareness through training rather than prompting alone. VisRep~\citep{khan2024visrep} and DWIM~\citep{ke2025dwim} both use feedback from tool execution to teach LLMs how to reason with external modules. VisRep performs self-training, implemented as supervised fine-tuning, from program execution feedback, while DWIM uses discrepancy-aware workflow generation and instruct-masking fine-tuning to help the model understand tool functionalities, detect potential errors, and revise decisions.

A complementary line constrains tool use through structured executable reasoning. LEFT~\citep{hsu2023left}, NAVER~\citep{cai2025naver}, SYNAPSE~\citep{modak2025synapse}, ViUniT~\citep{panagopoulou2025viunit}, VADAR~\citep{marsili2025visual_VARAR}, TANGO~\citep{Ziliotto_2025_CVPR_TANGO}, and PMA in CityEQA~\citep{zhao-etal-2025-cityeqa_pma} all move beyond unconstrained tool invocation by imposing executable structures over the reasoning process. LEFT reasons over LLM-proposed visual concepts with a differentiable logic executor, while NAVER and SYNAPSE integrate LLM-guided logic with structured tool invocation and verification. ViUniT validates visual programs using visual unit tests. VADAR dynamically constructs Pythonic APIs and uses a test agent to validate generated functions; TANGO composes executable programs from navigation and exploration primitives; and PMA adopts a hierarchical Planner--Manager--Actor architecture with an object-centric cognitive map for long-horizon visual reasoning.

\subsubsection{Synthesis and Outlook}
Overall, this class of methods reflects a shift from static prompting to interactive, interpretable, and dynamically evolving tool use, where LLMs are not just passive planners, but increasingly learn to reason with and through tools~\citep{ke2025dwim}. These advances lay the groundwork for future multimodal agents with stronger compositionality, reliability, and generalization. 

However, these methods have a critical bottleneck in that there is an unbounded number of interpretations from a finite set of visual building blocks. When LLMs operate solely on textual descriptions converted from visual inputs, their abstraction capacity becomes a major limitation. If the abstraction is inaccurate, misaligned, or fails to capture essential visual cues, crucial information may be lost during the conversion. This impairs the system’s ability to solve the task correctly, as important visual semantics are never incorporated into the reasoning process.

\begin{table*}[t]
\centering
\small
\setlength{\tabcolsep}{3.0pt}
\renewcommand{\arraystretch}{1.13}
\caption{
Taxonomy of representative tool-enhanced methods in Stage II and Stage III.
}
\label{tab:stage2_stage3_methods}

\resizebox{0.98\textwidth}{!}{%
\begin{tabular}{l c c c c  c c c  c}
\toprule
\textbf{Model}
& \textbf{Venue}
& \textbf{Stage}
& \textbf{Planner}
& \textbf{Tool Interface}
& \textbf{Control}
& \textbf{State / Feedback}
& \textbf{PnP Scope}
& \textbf{Core Training} \\
\midrule

\multicolumn{9}{l}{\textit{Stage II: Tool-enhanced LLMs}} \\
\midrule

Visual ChatGPT~\cite{wu2023visualchatgpt}
& arXiv 2023
& II
& LLM
& Workflow
& One-pass
& None
& TF wrapper
& Training-free \\

GPT4Tools~\cite{yang2023gpt4tools}
& NeurIPS 2023
& II
& LLM
& Tool call
& One-pass
& None
& Trained tool-user
& SFT \\

HuggingGPT~\cite{shen2023hugginggpt}
& NeurIPS 2023
& II
& LLM
& Workflow
& One-pass
& None
& TF wrapper
& Training-free \\

Chameleon~\cite{lu2023chameleon}
& NeurIPS 2023
& II
& LLM
& Workflow
& One-pass
& None
& TF wrapper
& Training-free \\

VisProg~\cite{gupta2023visprog}
& CVPR 2023
& II
& LLM
& Program
& One-pass
& None
& TF wrapper
& Training-free \\

ViperGPT~\cite{suris2023vipergpt}
& ICCV 2023
& II
& LLM
& Program
& One-pass
& None
& TF wrapper
& Training-free \\

MM-REACT~\cite{yang2023mmreact}
& arXiv 2023
& II
& LLM
& Tool call
& Iterative
& Text obs.
& TF wrapper
& Training-free \\

InternGPT~\cite{liu2023interngpt}
& arXiv 2023
& II
& LLM
& Tool call
& Iterative
& Text obs.
& TF wrapper
& Training-free \\

LEFT~\cite{hsu2023left}
& NeurIPS 2023
& II
& LLM
& Program
& One-pass
& Ver.
& Trained reasoner
& SFT \\

CLOVA~\cite{gao2024clova}
& CVPR 2024
& II
& LLM
& Program
& Iterative
& Mem.+Ver.
& Trained tool-user
& SFT \\

VisRep~\cite{khan2024visrep}
& CVPR 2024
& II
& LLM
& Program
& One-pass
& None
& Trained reasoner
& SFT \\

HYDRA~\cite{ke2024hydra}
& ECCV 2024
& II
& LLM
& Workflow
& Iterative
& Mem.+Ver.
& Trained policy
& RL \\

CRAFT~\cite{yuan2024craft}
& ICLR 2024
& II
& LLM
& Tool call
& One-pass
& None
& TF wrapper
& Training-free \\

VIoTGPT~\cite{zhong2025viotgpt}
& AAAI 2025
& II
& LLM
& Tool call
& One-pass
& None
& Trained tool-user
& SFT \\

ContextualCoder~\cite{shen2025contextualcoder}
& TMM 2025
& II
& LLM
& Program
& One-pass
& None
& TF wrapper
& Training-free \\

ViuniT~\cite{panagopoulou2025viunit}
& CVPR 2025
& II
& LLM
& Program
& One-pass
& Ver.
& TF wrapper
& Training-free \\

NAVER~\cite{cai2025naver}
& ICCV 2025
& II
& LLM
& Program
& Iterative
& Mem.+Ver.
& TF wrapper
& Training-free \\

SYNAPSE~\cite{modak2025synapse}
& AAAI 2025
& II
& LLM
& Program
& One-pass
& Ver.
& TF wrapper
& Training-free \\

DWIM~\cite{ke2025dwim}
& ICCV 2025
& II
& LLM
& Workflow
& Iterative
& Ver.
& Trained tool-user
& SFT \\

VADAR~\cite{marsili2025visual_VARAR}
& CVPR 2025
& II
& LLM
& Program
& Iterative
& Ver.
& TF wrapper
& Training-free \\

TANGO~\cite{Ziliotto_2025_CVPR_TANGO}
& CVPR 2025
& II
& LLM
& Program
& One-pass
& None
& TF wrapper
& Training-free \\

PMA~\cite{zhao-etal-2025-cityeqa_pma}
& EMNLP 2025
& II
& LLM
& Workflow
& Iterative
& Mem.
& TF wrapper
& Training-free \\

\midrule

\multicolumn{9}{l}{\textit{Stage III: Language-Mediated Grounding Action Control}} \\
\midrule

Image-of-Thought~\cite{zhou2024imageofthought}
& arXiv 2024
& III
& VLM
& Workflow
& One-pass
& None
& TF wrapper
& Training-free \\

P2G~\cite{chen2024p2g}
& arXiv 2024
& III
& VLM
& Tool call
& One-pass
& None
& TF wrapper
& Training-free \\

LLaVA-Plus~\cite{liu2024llavaplus}
& ECCV 2024
& III
& VLM
& Tool call
& One-pass
& None
& Trained tool-user
& SFT \\

Syn~\cite{li2024syn}
& CVPR 2024
& III
& VLM
& Workflow
& One-pass
& None
& Trained reasoner
& SFT \\

LATTE~\cite{ma2025latte}
& EMNLP 2025
& III
& VLM
& Tool call
& Iterative
& None
& Trained tool-user
& SFT \\

\midrule
\multicolumn{9}{l}{\textit{Stage III: Embedding-Mediated Grounding Action Control}} \\
\midrule

VITRON~\cite{fei2024vitron}
& NeurIPS 2024
& III
& VLM
& Latent routing
& One-pass
& None
& Latent-coupled
& SFT \\

VisionLLM v2~\cite{li2024visionllmv2}
& NeurIPS 2024
& III
& VLM
& Latent routing
& One-pass
& None
& Latent-coupled
& SFT \\

\midrule
\multicolumn{9}{l}{\textit{Stage III: Visual Feedback }} \\
\midrule

Wiki-PRF~\cite{gu2026knowledge_wikiprf}
& NeurIPS 2025
& III
& VLM
& Tool call
& One-pass
& Vis.
& Trained policy
& RL \\

VToolR1~\cite{wu2025vtool}
& ICLR 2026
& III
& VLM
& Tool call
& Iterative
& Vis.
& Trained policy
& RL \\

GCA~\cite{Chen_2026_CVPR_GCA}
& CVPR 2026
& III
& VLM
& Program
& Iterative
& Mem.+Vis.
& TF wrapper
& Training-free \\

VIREO~\cite{cheng2024vireo}
& EMNLP 2024
& III
& VLM
& Workflow
& Iterative
& Mem.+Vis.
& Trained plug-in
& SFT \\

Visual Sketchpad~\cite{hu2024visualsketchpad}
& NeurIPS 2024
& III
& VLM
& Program
& Iterative
& Mem.+Vis.
& TF wrapper
& Training-free \\

Creative Agents~\cite{cai2025creativeagents}
& UAI 2025
& III
& VLM
& Workflow
& One-pass
& Vis.
& TF wrapper
& Training-free \\

OpenThinkIMG~\cite{su2025openthinkimg}
& arXiv 2025
& III
& VLM
& Tool call
& Iterative
& Mem.+Vis.
& Trained policy
& SFT+RL \\

MACT~\cite{Yu_2026_CVPR_MACT}
& CVPR 2026
& III
& VLM
& Workflow
& Iterative
& Mem.+Vis.+Ver.
& Multi-agent plug-in
& SFT \\

\bottomrule
\end{tabular}%
}

\vspace{2pt}
\footnotesize{
\textbf{Planner} denotes the central module that selects, invokes, or coordinates tools.
\textbf{Tool Interface} specifies whether tools are invoked through direct tool calls, executable programs, multi-step workflows, or latent routing.
\textbf{Control} distinguishes single-pass execution from iterative replanning using intermediate observations.
\textbf{State / Feedback} summarizes whether the agent maintains memory (\emph{Mem.}), receives visual feedback (\emph{Vis.}), uses textual observations (\emph{Text obs.}), or performs verification (\emph{Ver.}).
\textbf{PnP Scope} distinguishes training-free wrappers (\emph{TF wrapper}), trained tool users, trained plug-in reasoners, trained tool-use policies, latent-coupled architectures, and multi-agent plug-ins.
\textbf{Core Training} refers only to the planner, reasoner, or tool-use policy, excluding training of external tools or low-level modules.
}
\end{table*}

\subsection{Stage III: Tool-Enhanced Vision-Language Models}
\label{sec:toolvlm}

Tool-enhanced VLMs extend tool-enhanced LLMs by replacing the underlying LLM with a VLM, enabling more direct and flexible visual interaction, as illustrated in Figure~\ref{fig:stage2} and Table~\ref{tab:stage2_stage3_methods}. \textbf{Unlike tool-enhanced LLMs where planners rely on preprocessed visual information from tools, this framework allows planners to directly receive raw images as input, reducing information loss and improving efficiency}. Large-scale VLMs are dynamically augmented with external tools, enabling selective invocation of specialized systems (\eg, object detectors, OCR engines, segmenters, and image generators) via natural language interfaces or latent control signals. This architecture enhances performance on complex tasks and mitigates issues such as hallucination and weak visual grounding.

Tool-enhanced VLMs can be broadly classified based on the modality and explicitness of their tool interaction. The first category, referred to as language-mediated control, includes models that issue interpretable instructions, such as natural language prompts or symbolic commands, to activate tool functionalities. The second category, termed embedding-mediated integration, encompasses models that coordinate tool use through learned latent representations, embedding control logic directly into the architecture without relying on explicit symbolic mediation.

\subsubsection{Language-Mediated Grounding Action Control}
Models such as Image-of-Thought~\citep{zhou2024imageofthought}, P2G~\citep{chen2024p2g}, LLaVA-Plus~\citep{liu2024llavaplus}, Syn~\citep{li2024syn}, VIREO~\citep{cheng2024vireo}, and GCA~\citep{Chen_2026_CVPR_GCA} exemplify VLM-centered planning paradigms. These systems rely on the VLM to generate interpretable grounding actions, including natural-language prompts, symbolic commands, code-like programs, or formal constraints, which are then used to invoke relevant external tools for perception and reasoning.

For example, Image-of-Thought~\citep{zhou2024imageofthought} simulates human cognitive processes by planning sequences of visual operations like segmentation or cropping and combining the resulting visual rationales with textual reasoning. P2G~\citep{chen2024p2g} facilitates fine-grained visual understanding by allowing the VLM to dynamically request help from OCR or grounding agents, particularly for high-resolution and text-rich images. LLaVA-Plus~\citep{liu2024llavaplus} employs a modular design in which the VLM acts as a planner that generates ``Thought-Action-Value'' prompts to control a repository of pre-trained visual tools. VIREO~\citep{cheng2024vireo} and LATTE~\citep{ma2025latte} improve multi-step reasoning by automatically decomposing questions and invoking tools step-by-step to resolve sub-questions.

Furthermore, with the emergence of GPT-o1 and DeepSeek-R1, reinforcement learning-based approaches for tool selection and intermediate evidence filtering have gained traction. OpenThinkIMG~\citep{su2025openthinkimg} and VToolR1~\citep{wu2025vtool} optimize visual tool use for image-based reasoning, while Wiki-PRF~\citep{gu2026knowledge_wikiprf} extends this idea to retrieval-augmented visual reasoning by training a VLM to dynamically invoke visual tools, retrieve multimodal knowledge, and filter irrelevant evidence using reward signals.

\subsubsection{Embedding-Mediated Grounding Action Control}
By contrast, VITRON~\citep{fei2024vitron} and VisionLLM v2~\citep{li2024visionllmv2} exemplify models in the second category, where interaction with external modules occurs through learned embeddings rather than textual prompts. These systems embed communication pathways directly into their architecture. VITRON~\citep{fei2024vitron} employs a hybrid message-passing framework that combines discrete routing tokens with continuous embeddings to activate various back-end modules for tasks like image editing, segmentation, and video generation. VisionLLM v2~\citep{li2024visionllmv2} introduces a ``super link'' mechanism that routes task-specific information using learnable query embeddings bound to routing tokens, allowing seamless integration of decoders for diverse tasks such as object detection, pose estimation, and image synthesis.

\subsubsection{Visual Feedback in Tool-Enhanced VLMs}
A more general class of tool-enhanced approaches for compositional visual reasoning aims to incorporate feedback during multi-turn, iterative processes. The most common form of feedback is textual, often delivered through natural language messages or structured programming outputs. However, when VLMs are used for action grounding and receive raw image inputs directly, this enables the system to obtain visual feedback in the form of updated or modified images. This gives rise to a distinct framework in which the model interacts with external tools to operate on images and then processes the resulting visual input for further reasoning.

A related direction introduces agent-wise collaboration and self-correction into VLM-centered tool use. MACT~\citep{Yu_2026_CVPR_MACT} decomposes visual reasoning into planning, execution, judgment, and answer agents, where the judgment agent verifies intermediate results and redirects earlier agents for revision, enabling adaptive test-time scaling and stronger factual grounding.

Models like Visual Sketchpad~\citep{hu2024visualsketchpad}, VToolR1~\citep{wu2025vtool}, Self-Imagine~\citep{Akter2024SelfImagine}, Creative Agents~\citep{cai2025creativeagents}, and GCA~\citep{Chen_2026_CVPR_GCA} extend the visual-feedback paradigm by using external tools to synthesize imagined, modified, or intermediate visual states during reasoning.


\subsubsection{Synthesis and Outlook}
While tool-enhanced VLMs demonstrate strong generalization, grounded reasoning, and modular interpretability, they also present distinct challenges. These include high computational demands, increased input complexity, reliance on external tool accuracy, and architectural overhead~\citep{yang2023gpt4tools, chen2024p2g, alayrac2022flamingo, li2024visionllmv2}. Future work should focus on improving integration robustness and efficiency—moving from prompt-based orchestration toward seamless architectural unification—to enable scalable and general-purpose multimodal agents~\citep{zhao2025cotvla}.
\subsection{Stage IV: Chain-of-Thought Reasoning VLMs}
\label{sec:deepthinkvlm}
\begin{figure*}[t]
\begin{center}
  \includegraphics[width=0.95\linewidth]{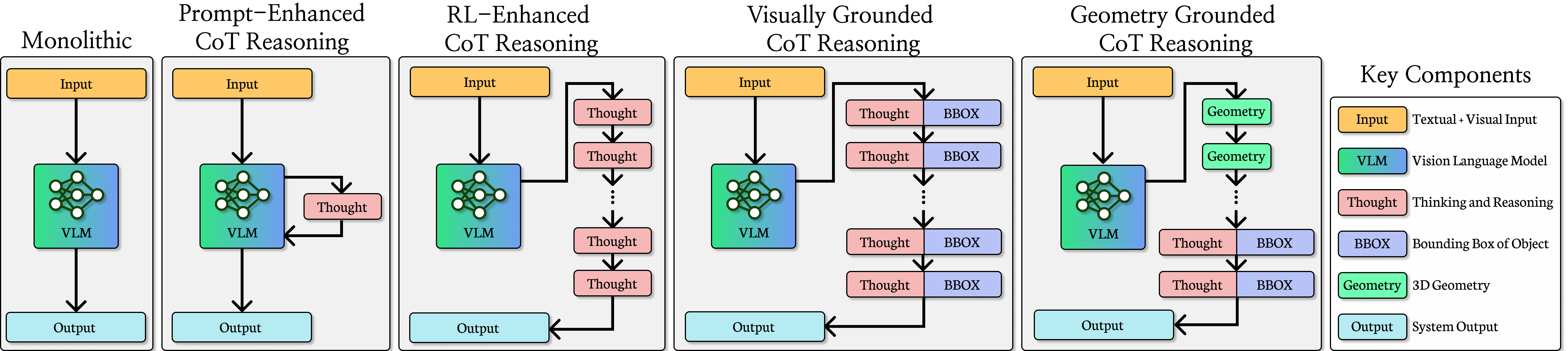} 
\end{center}
\vspace{-1em}
\caption{Monolithic versus Chain-of-Thought (CoT) reasoning VLMs (Stage IV) pipeline.}
\label{fig:stage4}
\end{figure*}

\begin{table*}[t]
\centering
\scriptsize
\setlength{\tabcolsep}{2.5pt}
\renewcommand{\arraystretch}{1.13}
\caption{
Taxonomy of representative Chain-of-Thought reasoning VLMs in Stage IV.
}
\label{tab:stage4_cot_methods}

\resizebox{0.99\textwidth}{!}{%
\begin{tabular}{l c c c c c l}
\toprule
\textbf{Model}
& \textbf{Venue}
& \textbf{Subtype}
& \textbf{Trace Form}
& \textbf{Evidence}
& \textbf{Core Signal}
& \textbf{Main Scope} \\
\midrule

\multicolumn{7}{l}{\textit{Stage IV-A: Prompt-enhanced CoT Reasoning VLMs}} \\
\midrule

LLaVA-CoT~\citep{xu2025llavacotletvisionlanguage}
& ICCV 2025
& Prompt-CoT
& Text stages
& Img
& SFT
& General multimodal reasoning \\

CCoT~\citep{mitra2024ccot}
& CVPR 2024
& Prompt-CoT
& Scene graph
& ObjRel
& Prompt
& Compositional reasoning \\

\midrule
\multicolumn{7}{l}{\textit{Stage IV-B: RL-enhanced CoT Reasoning VLMs}} \\
\midrule

KARL~\citep{ma2026KARL}
& arXiv 2025
& RL-CoT
& Perception trace
& VisTrace
& SFT+RL
& Knowledge-intensive visual grounding \\

G1~\citep{chen2025g1}
& arXiv 2025
& RL-CoT
& Reasoning trajectory
& Img
& RL
& Bootstrapped perception and reasoning \\

Ground-R1~\citep{cao2025groundr1}
& arXiv 2025
& RL-CoT
& Grounded CoT
& VisTrace
& RL
& Grounded visual reasoning \\

VisRL~\citep{chen2025visrl}
& ICCV 2025
& RL-CoT
& Intention trace
& VisTrace
& RL
& Intention-driven perception \\

Reason-RFT~\citep{tan2025reasonrft}
& NeurIPS 2025
& RL-CoT
& Reasoning trajectory
& Img
& SFT+RL
& General visual reasoning \\

VLM-R1~\citep{shen2025vlmr1}
& arXiv 2025
& RL-CoT
& R1-style trace
& Img
& SFT+RL
& Stable R1-style VLM reasoning \\

Vision-R1~\citep{huang2025visionr1}
& ICLR 2026
& RL-CoT
& Long CoT
& Img
& SFT+RL
& Multimodal reasoning capability \\

\midrule
\multicolumn{7}{l}{\textit{Stage IV-C: Visually grounded CoT Reasoning VLMs}} \\
\midrule

PaLI-X-VPD~\citep{hu2024pali}
& CVPR 2024
& VG-CoT
& Program trace
& ToolTrace
& Distill
& Distilling tool reasoning into VLMs \\

VoCoT~\citep{li-etal-2025-vocot}
& NAACL 2025
& VG-CoT
& Region tuple
& Box
& SFT
& Visually grounded multi-step reasoning \\

LLaVA-Aurora~\citep{bigverdi2025LLAVA_AURORA}
& CVPR 2025
& VG-CoT
& Perception tokens
& Token
& SFT
& Fine-grained perception-enhanced reasoning \\

VISTAR~\citep{cheng2025VISTAR}
& CVPRW 2025
& VG-CoT
& Subtask trace
& VisTrace
& SFT
& Interpretable VQA reasoning \\

MM-GCoT~\citep{Wu_2026_CVPR_mmgcot}
& CVPR 2026
& VG-CoT
& Grounded CoT
& VisTrace
& SFT
& Answer-grounding consistency \\

CoReS~\citep{bao2024cores}
& ECCV 2024
& VG-CoT
& Seg.-logic chain
& Mask
& SFT
& Reasoning segmentation \\

\midrule
\multicolumn{7}{l}{\textit{Stage IV-D: Geometry- and Cross-view-grounded CoT Reasoning VLMs}} \\
\midrule

GeoThinker~\citep{GeoThinker}
& ICML 2026
& Geo-CoT
& Geometry trace
& Geo3D
& Distill
& Spatial reasoning with 3D priors \\

3DThinker~\citep{Chen_2026_CVPR_3dthinker}
& CVPR 2026
& Geo-CoT
& 3D latent trace
& Geo3D
& Distill
& 3D geometric imagination \\

VIEW2SPACE~\citep{ke2026view2space}
& ECCV 2026
& XView-CoT
& Multi-view CoT
& Box
& SFT
& Cross-view spatial reasoning \\

MindCube~\citep{wang2026mindcube}
& ICLR 2026
& XView-CoT
& Cognitive map
& ObjRel
& SFT
& Spatial mental modeling \\

\bottomrule
\end{tabular}%
}

\vspace{2pt}
\footnotesize{
\textbf{Subtype} denotes the main mechanism used to elicit or learn CoT reasoning:
Prompt-CoT, RL-CoT, visually grounded CoT (VG-CoT), geometry-distilled CoT (Geo-CoT), cross-view CoT (XView-CoT), and unified end-to-end CoT (Unified-CoT).
\textbf{Trace Form} summarizes the intermediate reasoning representation, such as textual stages, scene graphs, visual programs, grounded tuples, segmentation chains, geometry traces, 3D latent traces, cognitive maps, or unified understanding-thinking-answering traces.
\textbf{Evidence} uses controlled labels:
Img = image-level evidence;
ObjRel = object-relation structure;
Box = bounding boxes or region references;
Mask = segmentation masks;
Token = learned perception tokens;
ToolTrace = tool-generated traces;
VisTrace = explicit visual evidence or grounded observations;
Geo3D = geometric or 3D priors;
MultiView = sparse or cross-view observations;
HiRes = high-resolution visual evidence.
\textbf{Core Signal} denotes the dominant learning or supervision signal:
Prompt, SFT, Distill, RL, or SFT+RL.
}
\end{table*}

Recent developments in vision-language models (VLMs) have led to the emergence of a new generation of chain-of-thought (CoT) reasoning-centric, end-to-end architectures as illustrated in Table~\ref{tab:stage4_cot_methods} and Figure~\ref{fig:stage4}. Unlike prompt-enhanced or tool-enhanced systems that rely on external modules and prompt chaining, these models aim to integrate perception and reasoning into a single forward pass, leveraging pretraining and fine-tuning strategies inspired by large-scale LLM developments, such as GPT-o1~\citep{jaech2024openaigpto1} and DeepSeek-R1~\citep{guo2025deepseekr1}. This section categorizes these models into four evolving trends: 1) Prompt-enhanced CoT reasoning VLMs; 2) RL-enhanced CoT reasoning VLMs; 3) Visually grounded CoT reasoning VLMs; 4) Geometry- and Cross-view-grounded CoT Reasoning VLMs. 

\subsubsection{Prompt-Enhanced CoT Reasoning VLMs} 
Prompt-enhanced CoT reasoning VLMs simulate multi-step reasoning behavior within a standard vision-language pipeline by leveraging in-context prompting or explicit stage-wise instruction formats. These models often frame the problem through intermediate representations or structured sub-tasks that guide reasoning while avoiding external modules or extra learning. A representative example is LLaVA-CoT~\citep{xu2025llavacotletvisionlanguage}, which breaks its output into structured reasoning stages, including summarization, captioning, intermediate deduction, and final conclusion. Similarly, CCoT~\citep{mitra2024ccot} enhances compositionality by instructing the model to generate scene graphs from the input, which then guide the reasoning process through object-centric abstractions. 
GPT-o1-style prompting strategies extend this approach by defining explicit intermediate stages that structure the reasoning process in a predetermined manner. While these methods demonstrate improved performance and explainability through fixed structures, their reasoning capacity remains constrained by limited adaptability and generalization across diverse tasks.

\subsubsection{RL-Enhanced CoT Reasoning VLMs} 
Inspired by proprietary reasoning models such as GPT-o1~\citep{jaech2024openaigpto1} and DeepSeek-R1~\citep{guo2025deepseekr1}, RL-enhanced CoT reasoning VLMs internalize systematic reasoning through combined supervised fine-tuning and reinforcement learning. These models typically adopt a two-phase learning framework: an initial supervised fine-tuning stage to encode reasoning patterns, followed by reinforcement learning or curriculum-based refinement to improve generalization and robustness. For example, Reason-RFT~\citep{tan2025reasonrft}, Vision-R1~\citep{huang2025visionr1}, VLM-R1~\citep{shen2025vlmr1} and VisRL~\citep{chen2025visrl} leverage Group Relative Policy Optimization (GRPO)~\citep{shao2024deepseekgrpo} to balance logical coherence with perceptual grounding during reasoning. KARL~\citep{ma2026KARL} introduces a cognition-perception synergy where reasoning accuracy and visual comprehension mutually enhance each other. G1~\citep{chen2025g1} adapts these strategies for visually interactive settings, enabling perception-grounded decision-making across complex tasks. Ground-R1~\citep{cao2025groundr1} introduces interpretable reasoning by decoupling visual evidence grounding from answer generation, trained end-to-end using only question-answer pairs without external annotations. 

The two-phase learning framework aligns with findings that RL can encourage VLMs to actively search for correct answers~\citep{cao2025groundr1, guo2025deepseekr1}. From an optimization standpoint, the performance improvements from RL (\eg, RLHF) stem from how it reshapes the model's output distribution. In distributional terms, the SFT objective corresponds to trajectory-level distribution matching using forward KL divergence ($KL(P_{data} \Vert Q_{\theta})$), encouraging the model $\theta$ to broadly match the empirical distribution of the training data. While maximizing this distributional coverage ensures diversity, it also assigns probability mass to plausible but suboptimal or irrelevant outputs. In contrast, RL shifts the objective toward mode-seeking via reverse KL divergence ($KL(Q_{\theta} \Vert P_{data})$), training the model to concentrate specifically on generating outputs that are consistently correct or optimal. Although this reduces behavioral diversity, it yields more coherent and accurate responses, which is particularly beneficial in multi-step reasoning tasks where correctness and logical consistency outweigh broad coverage~\citep{xiao2025connection, wang2023beyond, chu2025sft}.

Although these models represent a significant shift toward more autonomous and cognitively aligned reasoning architectures, they often require substantial computational resources and careful curriculum design to achieve robust performance. While both RL-enhanced and prompt-enhanced CoT reasoning VLMs provide interpretable solutions by exposing structured reasoning steps, they still depend on implicit reasoning paths and typically lack explicit intermediate supervision or visual grounding.

\subsubsection{Visually Grounded CoT Reasoning VLMs} 
Visually grounded CoT reasoning VLMs explicitly couple reasoning steps with visual grounding information, producing interpretable intermediate outputs such as bounding boxes, object references, or spatial annotations alongside final answers. This class of models benefits from training data that aligns structured reasoning with grounded visual supervision, either synthesized via tool-based pipelines or adapted from existing multimodal datasets ~\citep{hudson2019gqa, marino2019okvqa, schwenk2022aokvqa}. 

A representative example is PaLI-X-VPD~\citep{hu2024pali}, which introduces visual program distillation to inject coherent, tool-generated chain-of-thought reasoning traces into the training process. Building on the idea of explicit object grounding, VoCoT~\citep{li-etal-2025-vocot}, LLaVA-Aurora~\citep{bigverdi2025LLAVA_AURORA} formulates each reasoning step as a tuple combining a visual region, its spatial coordinates, and a semantic description, enabling object-centric, multi-step inference. MM-GCoT~\citep{Wu_2026_CVPR_mmgcot} and VISTAR~\citep{cheng2025VISTAR} contribute a dedicated dataset designed to enforce answer-grounding consistency, encouraging VLMs to align each reasoning step with grounded visual evidence. CoReS~\citep{bao2024cores} advances this line by combining segmentation and logical inference through interleaved dual chains, improving compositional and fine-grained visual reasoning.

Despite their interpretability and stronger grounding, these models often depend on costly training pipelines and high-quality grounded traces, which may limit scalability and generalization in less structured environments.

\subsubsection{Geometry- and Cross-view-grounded CoT Reasoning VLMs}

Geometry-distilled and cross-view CoT reasoning VLMs extend Stage IV by injecting spatially structured evidence into the internal reasoning process of VLMs. 
Unlike visually grounded CoT methods that mainly supervise intermediate steps with explicit 2D evidence such as boxes, masks, or object references, this direction distills or constructs geometric priors from depth cues, 3D representations, viewpoint transformations, and limited-view observations. 
As a result, reasoning is guided not only by visible regions or textual rationales, but also by implicit spatial structures that encode depth, layout, occlusion, viewpoint consistency, and cross-view correspondences.

GeoThinker~\citep{GeoThinker} actively integrates geometric evidence for spatial inference, while 3DThinker~\citep{Chen_2026_CVPR_3dthinker} grounds reasoning in 3D geometric imagination from limited views. 
MindCube~\citep{wang2026mindcube} and VIEW2SPACE~\citep{ke2026view2space} further extend this paradigm to cross-view spatial reasoning, where grounded traces may take the form of cognitive maps, viewpoint-aware evidence, or multi-view spatial correspondences. 
Together, these methods move CoT VLMs from language- or region-grounded reasoning toward geometry-aware spatial reasoning, where 3D priors and cross-view evidence serve as implicit grounding signals for compositional inference. 
However, their reliance on geometric supervision, 3D foundation models, simulation data, or limited-view annotations may constrain scalability, and their latent spatial reasoning process is often less interpretable than explicit grounding traces.

\subsubsection{Synthesis and Outlook}

CoT reasoning VLMs bridge perception and inference by exposing intermediate reasoning steps. 
Prompt-enhanced and RL-enhanced methods improve the structure and reliability of reasoning, while visually grounded CoT models align intermediate steps with explicit 2D evidence such as boxes, masks, or object references. 
More recent geometry-distilled approaches further push this paradigm by injecting 3D grounding signals into VLMs, encouraging models to produce CoT traces that encode depth, layout, viewpoint consistency, and latent spatial structure. 
The next frontier is therefore to develop CoT VLMs whose reasoning is not only linguistic and visually grounded, but also geometrically aware. 
However, the predominantly single-pass nature of current CoT architectures still limits iterative evidence inspection, self-correction, and refinement for complex visual reasoning tasks.

\subsection{Stage V: Unified Agentic Vision-Language Models}
\label{sec:agentvlm}

\begin{figure*}[t]
\begin{center}
  \includegraphics[width=0.9\linewidth]{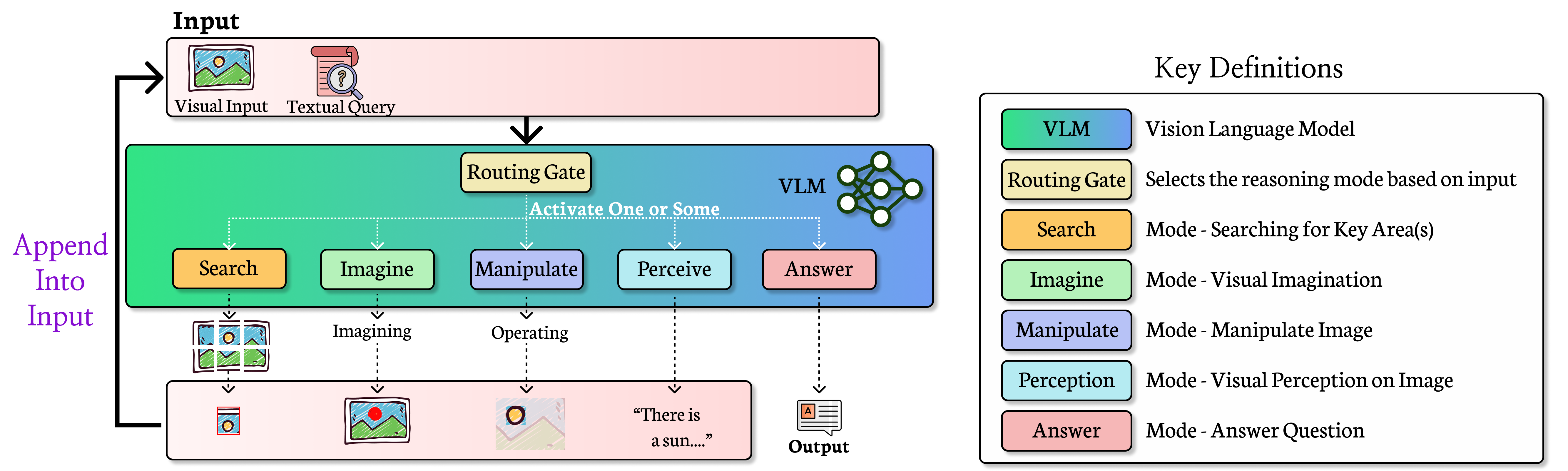} 
\end{center}
\vspace{-1em}
\caption{Illustration of unified agentic VLM (Stage V) inference. The model iteratively refines its understanding by appending intermediate outputs to the input and dynamically invoking internal capabilities to resolve complex visual reasoning tasks.}
\label{fig:stage5}
\end{figure*}

\begin{table*}[t]
\centering
\small
\setlength{\tabcolsep}{3.0pt}
\renewcommand{\arraystretch}{1.12}
\caption{
Representative unified agentic VLMs in Stage V.
}
\label{tab:stage5_agentic_vlms}

\resizebox{0.98\textwidth}{!}{%
\begin{tabular}{l l l l l c l l}
\toprule
\textbf{Model}
& \textbf{Venue}
& \textbf{Subtype}
& \textbf{Visual State}
& \textbf{Action / Mechanism}
& \textbf{Memory}
& \textbf{Feedback}
& \textbf{Training} \\
\midrule

\multicolumn{8}{l}{\textit{Explicit visual exploration and goal-driven evidence acquisition}} \\
\midrule
SEAL~\citep{wu2024vstar}
& CVPR 2024
& Exploration
& Retrieved patches
& Guided visual search
& \cmark
& Visual
& Training-free \\

DC2~\citep{wang2025dc2}
& AAAI 2025
& Exploration
& Image patches
& Divide-conquer-combine
& \cmark
& Visual/Text
& Training-free \\

ZoomEye~\citep{shen2024zoomeye}
& arXiv 2024
& Exploration
& Patch tree
& Confidence-guided zooming
& \cmark
& Confidence
& Training-free \\

FAST~\citep{sunfast}
& ICLR 2025
& Exploration
& Evidence chain
& Fast-slow visual reasoning
& \cmark
& Visual
& Training-free \\

CogCoM~\citep{qi2024cogcom}
& ICLR 2025
& Exploration
& Manipulation trace
& Chain-of-manipulations
& \cmark
& Visual
& SFT \\

GeReA~\citep{ma2024gerea}
& arXiv 2024
& Exploration
& Question-aware captions
& Prompt-caption evidence
& \xmark
& Text
& Training-free \\

Insight-V~\citep{dong2025insightv}
& CVPR 2025
& Exploration
& Long visual trace
& Long-chain visual reasoning
& \cmark
& Visual/Text
& SFT \\

Argus~\citep{man2025argus}
& CVPR 2025
& Exploration
& Grounded CoT
& Vision-centric reasoning
& \cmark
& Visual
& SFT \\

VisCoT~\citep{shao2024visualcot}
& NeurIPS 2024
& Exploration
& Visual CoT
& Step-wise visual reasoning
& \xmark
& Visual/Text
& SFT \\

ViGoRL~\citep{sarch2025ViGoRL}
& NeurIPS 2026
& Exploration
& Grounded states
& Grounded reinforcement learning
& \cmark
& Reward
& RL \\

SIFThinker~\citep{chen2025sifthinker}
& AAAI 2026
& Exploration
& Focused image regions
& Spatially-aware image focus
& \cmark
& Visual
& SFT \\

DeepEyes~\citep{zheng2026deepeyes}
& ICLR 2026
& Exploration
& Image-based states
& Thinking with images
& \cmark
& Reward/Visual
& RL \\

\midrule
\multicolumn{8}{l}{\textit{Latent visual-state reasoning and world-model-like imagination}} \\
\midrule
Mirage~\citep{Yang2025Mirage}
& arXiv 2025
& Latent imagination
& Latent visual tokens
& Machine mental imagery
& \cmark
& Latent
& SFT \\

FOREWARN~\citep{wu2025FOREWARN}
& RSS 2025
& Latent visual state
& Latent world state
& World-model dynamics
& \cmark
& Latent
& SFT \\

Astra~\citep{zhu2026thinkingimage_astra}
& arXiv 2026
& World imagination
& Novel-view observations
& World-simulator imagination
& \cmark
& Simulated visual
& RL \\

\bottomrule
\end{tabular}%
}
\end{table*}

Unified agentic VLMs represent a recent shift in multimodal reasoning, wherein the model actively plans, adapts, imagines, and executes a sequence of decisions to solve complex visual tasks~\citep{sunfast, wu2024vstar}. Different from prompt-enhanced or tool-augmented systems, where language models issue static queries or deterministic tool calls, unified agentic VLMs are designed to autonomously decompose tasks, selectively gather visual evidence, and iteratively refine their reasoning process. These models exhibit a form of learned deliberation: they can identify uncertainty, determine what additional visual evidence is needed, and justify their answers through explicit reasoning and action trajectories, as illustrated in Figure~\ref{fig:stage5} and Table~\ref{tab:stage5_agentic_vlms}.

These models~\citep{qi2024cogcom, wang2025dc2} often operate within a multi-step feedback loop that integrates perception, decision-making, execution, and memory. 
A typical architecture consists of a controller, usually an LLM, VLM, or learned policy model, together with visual engines or internal visual operations, such as captioning, segmentation, detection, grounding, zooming, OCR, or region retrieval, and an intermediate memory or context accumulation module. 
The reasoning process can be conducted explicitly, through structured chains of visual operations, or implicitly, through learned state representations. 
Accordingly, Stage V methods can be broadly organized into two directions. 
The first emphasizes explicit visual exploration, where models discover informative regions, zoom into image patches, perform visual operations, and accumulate evidence in memory to better exploit visual information already present in the input. 
The second emphasizes visual-state-augmented reasoning, where models maintain latent visual states, imagined representations, or world-model-like dynamics to support internal imagination, belief refinement, and reasoning over information that is not directly observed.


\subsubsection{Automatic Discovery of Informative Regions and Goal-Driven Exploration}
A subset of unified agentic VLMs demonstrates the ability to automatically identify, search, and discover informative regions within an image to iteratively solve subtasks and ultimately complete the overall task. These models exemplify active visual reasoning by dynamically allocating attention and modifying inputs based on task demands.

Representative systems illustrate this paradigm in diverse ways. SEAL~\citep{wu2024vstar} incorporates a visual working memory that stores the question, global image, target regions, and retrieved patches. When the initial image context proves insufficient, the model identifies missing visual entities, performs targeted search and cropping, and revisits the question with enriched inputs. ZoomEye~\citep{shen2024zoomeye} treats the image as a hierarchical patch tree and uses confidence scores from an MLLM to guide zoom-in actions and determine when to stop based on answer certainty. FAST~\citep{sunfast} adopts a dual-system architecture that switches between fast and slow reasoning modes. When greater precision is needed, it activates built-in visual operations, such as segmentation and proposal generation, and aggregates the resulting evidence through a chain-of-evidence structure.

Other methods, such as CogCoM~\citep{qi2024cogcom}, VisCoT~\citep{shao2024visualcot}, Insight-V~\citep{dong2025insightv}, GeReA~\citep{ma2024gerea}, ViGoRL~\citep{sarch2025ViGoRL}, SIFThinker~\citep{chen2025sifthinker}, Argus~\citep{man2025argus} and DeepEyes~\citep{zheng2026deepeyes} guide an LLM to generate a sequence of visual manipulation steps—such as grounding, zooming, OCR, and counting—to incrementally acquire task-relevant information. 
DC2~\citep{wang2025dc2} proposes an alternative strategy by dividing high-resolution images into patches, summarizing them individually, and using a visual memory mechanism to selectively retrieve relevant regions at inference time.

\subsubsection{Latent Imagination and World-Model-Augmented Reasoning}
Another class of unified agentic VLMs augments reasoning with latent or imagined visual states rather than only inspecting visual evidence already present in the input. 
These methods use latent visual tokens, hidden visual representations, world-model dynamics, or simulator-generated observations to support internal imagination and reasoning beyond directly observed visual content.

Some models focus on textual imagination within the latent space. For example, Mirage~\citep{Yang2025Mirage} introduces a machine mental imagery framework that augments VLM decoding with latent visual tokens alongside textual outputs. Rather than generating pixel-level images, the model performs ``visual thinking'' by extending hidden state trajectories with multimodal tokens, maintaining a multimodal reasoning flow entirely in latent space. Another notable approach, FOREWARN~\citep{wu2025FOREWARN}, adapts the LLaMA-3.2-11B-Vision-Instruct model by replacing its image tokenization module with a world model encoder and latent dynamics component. This design allows the model to internally simulate visual states during reasoning, enabling it to perform latent imagination without relying on external rendering. Moving from latent simulation to explicit simulator-mediated imagination, Astra~\citep{zhu2026thinkingimage_astra} actively requests novel-view observations from an integrated world simulator. It uses a world simulator to produce novel-view visual observations, allowing the model to reason over unobserved spatial layouts and cross-view evidence beyond the original input.


\subsubsection{Synthesis and Outlook}
Unified agentic VLMs enhance compositionality, adapt to ambiguous inputs, and support interpretability through intermediate visual states, visual memories, or visual artifacts~\citep{wang2025dc2, sunfast, qi2024cogcom, zheng2026deepeyes}. Many show strong performance on high-resolution and compositional benchmarks such as V*Bench~\citep{wu2024vstar}, GQA~\citep{hudson2019gqa}, ReasonSeg~\citep{lai2024ReasonSeg}, and MM-Vet~\citep{mmvetyu2024}. They also demonstrate robustness when paired with relatively small LLMs, offering an efficient alternative to large proprietary models~\citep{zhan2025griffonr}.

Nonetheless, these systems face challenges. Many rely on handcrafted prompts, special tokens, or rule-based triggers, which limit scalability~\citep{liu2024llavaplus, bao2024cores}. Multi-step pipelines often incur higher computational costs and may propagate errors through sequential stages. Visual operations such as cropping or segmentation are sometimes imprecise, especially for small or occluded objects~\citep{ke2025dwim, shen2024zoomeye, gao2024clova}. Moreover, the lack of unified supervision across reasoning steps and final outputs hinders end-to-end optimization~\citep{hu2024pali, cao2025groundr1}.

In summary, agentic vision-language models represent a promising direction that combines planning, perception, and reasoning within an integrated framework. They bridge the gap between passive captioning and interactive visual understanding. Future work may focus on improving their decision policies, unifying learning objectives, and integrating them with broader multimodal toolchains to support general-purpose visual intelligence.
\section{Benchmark and Evaluation}
\label{sec:benmark_evaluation}
Evaluation and benchmarking play a pivotal role in assessing the progress and diagnosing the limitations of visual reasoning systems. A wide range of datasets and metrics have been developed to probe different aspects of reasoning, minimize dataset bias, and analyze model capabilities with fine-grained annotations.

\subsection{Types of Benchmarks and Datasets}
\begin{figure*}[t]
\begin{center}
  \includegraphics[width=0.95\linewidth]{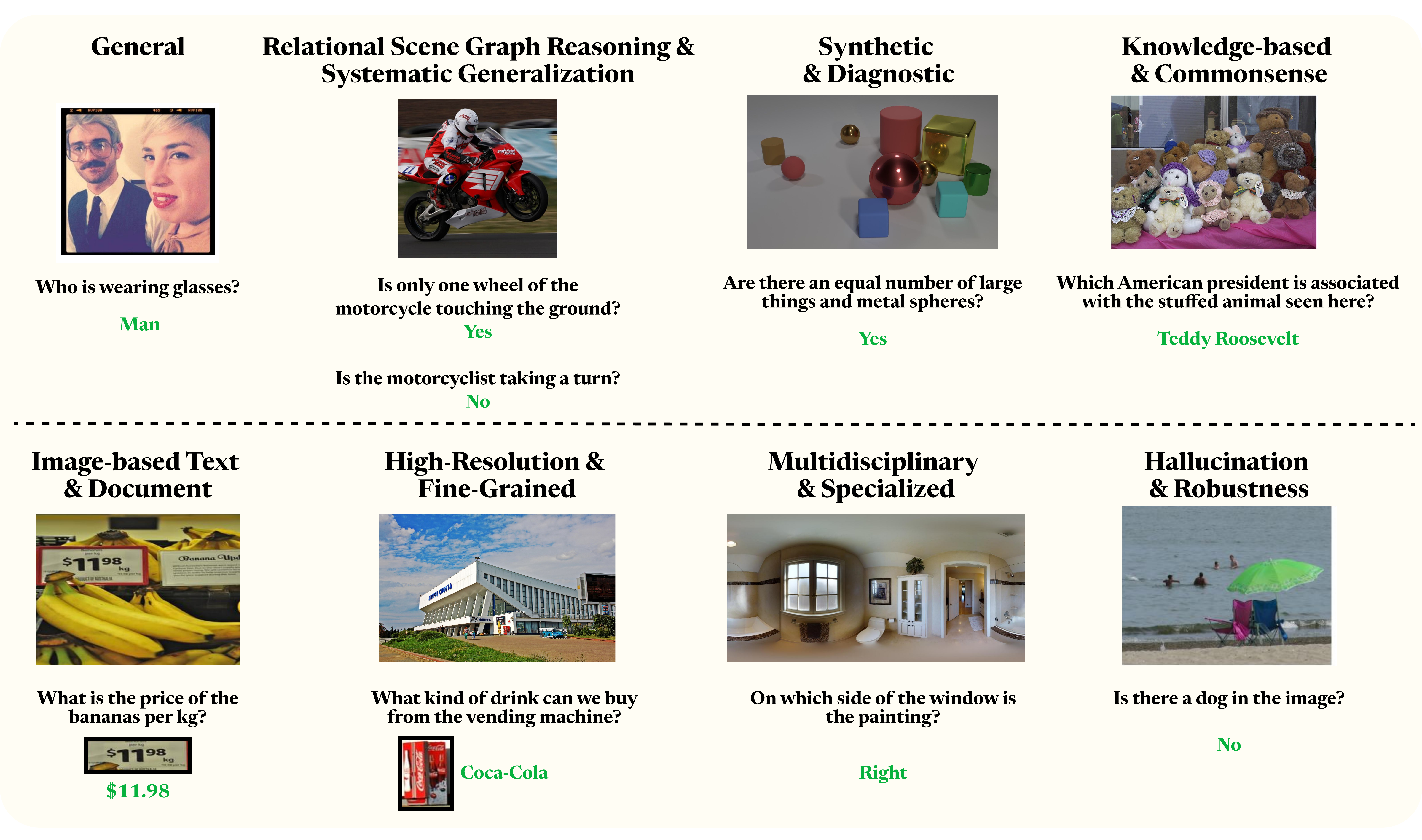} 
\end{center}
\vspace{-1.5em}
\caption{Examples of benchmarks and datasets.}
\label{fig:dataexp}
\vspace{-1em}
\end{figure*}

This subsection presents examples of different types of benchmarks and datasets, as illustrated in Figure~\ref{fig:dataexp}.

\textbf{General Visual Question Answering.}
General VQA benchmarks aim to evaluate a model’s ability to answer natural-language questions based on real-world images. These datasets focus on broad coverage of visual concepts and diverse question types (\eg, object presence, color, location, and actions), usually with human-generated questions. Prominent examples include VQA-v1~\citep{antol2015vqa}, VQA-v2~\citep{goyal2017vqa02}, COCO-QA~\citep{lin2014microsoftcocodataset}, Visual Genome~\citep{krishna2017visualgenome}, and Visual7W~\citep{zhu2016visual7w}. These datasets focus broadly on object presence, attributes, relationships, and basic visual comprehension. Variants such as VQA-CP~\citep{agrawal2018VQA-CP} specifically address biases through carefully restructured training and testing splits.
In short, general VQA datasets are mostly well-suited for testing perception-based understanding grounded in image content alone.

\textbf{Relational Scene Graph Reasoning and Systematic Generalization}
Relational scene graph reasoning and systematic generalization datasets are designed to evaluate a model’s ability to reason over novel combinations of familiar visual concepts, or interpret high-level semantic concepts across unseen environments. These tasks require models to perform multi-step inference, manipulate symbolic representations, and handle compositional semantics such as nested attributes, spatial relations, and logical structures. Often constructed using synthetic or structured environments, these benchmarks allow precise control over data generation and enable evaluation of reasoning fidelity and generalization capabilities under distribution shifts.

Systematic generalization is rigorously tested by datasets such as GQA~\citep{hudson2019gqa},
ReasonSeg~\citep{lai2024ReasonSeg}, TallyQA~\citep{acharya2019tallyqa}, NaturalBench~\citep{li2024naturalbench}, Cola~\citep{ray2023cola}, the Visual Abstractions Benchmark~\citep{hsu2025makes}, CREPE~\citep{ma2023crepe} and SugarCrepe~\citep{hsieh2023sugarcrepe}. These datasets explicitly target the models' compositional generalization, relational learning, and reasoning under uncertainty. In short, this category is essential for assessing structured reasoning, generalization to novel combinations, and step-wise inference fidelity.

\textbf{Synthetic and Diagnostic Evaluation.}
Synthetic and diagnostic benchmarks provide controlled environments to test specific reasoning sub-skills, such as attribute comparison, spatial relation tracking, and logical consistency. These datasets are often built with artificial scenes, minimizing irrelevant variability and allowing fine-grained analysis of reasoning behaviors. Diagnostic variants further introduce linguistic perturbations, attention supervision, or adversarial rephrasings to evaluate robustness and interpretability~\citep{kim2025visualsurvey, kabir2024comprehensive}.

Representative datasets include CLEVR~\citep{johnson2017clevr}, SHAPES~\citep{andreas2016SHAPES}, TaskMeAnything~\citep{zhang2024taskmeanything}, SVRT~\citep{Fleuret2011SVRT}, CLOSURE~\citep{Bahdanau2019CLOSUREAS}, CURI ~\citep{vedantam2021curi} and CLEVR variants (\eg, CLEVR-Ref+~\citep{liu2019Clevr-ref+}, CLEVR-XAI~\citep{arras2022clevrXAI}, QLEVR~\citep{li2022qlevr}, CLEVR-X~\citep{Salewski2022CLEVRX}, CLEVR-Dialog ~\citep{Kottur2019CLEVRDialog}). Diagnostic benchmarks like VQA-X~\citep{park2018VQAX}, VQA-HAT~\citep{das2017VQAHAT}, and VQA-Rephrasings~\citep{shah2019VQARephrasing} specifically target reasoning quality, explanation generation, human attention alignment, and linguistic robustness. Additionally, synthetic tests for abstract reasoning also fall under this category, notably, Raven's Progressive Matrices datasets (\eg, PGM~\citep{Santoro2018PGM}, RAVEN~\citep{Zhang2019RAVENAD}). Other examples include KANDINSKYPatterns~\citep{holzinger2021kandinskypatterns} and Bongard-LOGO~\citep{nie2020bongardlogo}, which focus on structural pattern recognition and analogy-based concept learning, respectively.

\textbf{Knowledge-Based and Commonsense Reasoning.}
Knowledge-based and commonsense benchmarks are developed to assess a model’s capacity to incorporate external knowledge (\eg, factual, cultural and commonsense) into the visual reasoning process. Unlike general VQA, these tasks require inference beyond observable pixels, leveraging external sources or prior training on world knowledge.

To this end, a number of datasets have been proposed to evaluate such capabilities. Examples include OK-VQA~\citep{marino2019okvqa}, A-OKVQA~\citep{schwenk2022aokvqa}, VCR~\citep{zellers2019recognition_VCR}, FVQA~\citep{wang2017fvqa}, and KB-VQA~\citep{wang2015KBVQA}, which require models to reason with factual or commonsense knowledge not present in the image. In addition, CRIC~\citep{Gao2019CRIC} further emphasizes compositional and grounded commonsense reasoning. These benchmarks are therefore crucial for evaluating whether a model can move beyond surface-level perception and engage in truly knowledge-intensive visual inference.

\textbf{Image-Based Text and Document Understanding.}
These benchmarks focus on evaluating a model’s ability to extract, read, and reason over textual information embedded in visual formats such as photographs, charts, and scanned documents. Tasks typically involve OCR-based understanding, document layout parsing, or diagram interpretation, requiring alignment between textual and spatial structures.
Representative datasets include TextVQA~\citep{kovaleva2020textvqa}, ST-VQA~\citep{biten2019STVQA}, DocVQA~\citep{mathew2021docvqa}, and AI2D~\citep{kembhavi2016diagramAI2D}.

\textbf{High-Resolution and Fine-Grained Perception.}
To explore vision-language models' capabilities in fine-grained perception, high-resolution benchmarks such as V*Bench~\citep{wu2024vstar}, HR-Bench (4K/8K)~\citep{wang2025dc2}, and LISA-Grounding~\citep{lai2024ReasonSeg} evaluate models' performance on detailed visual scenes and precise object recognition.

\textbf{Multidisciplinary and Specialized Domains.}
Benchmarks under this category extend visual reasoning into specialized domains or underrepresented user populations. They include medical diagnostics, accessibility for the visually impaired, cultural diversity, mathematical and chart reasoning, and embodied agents~\citep{lau2018VQARAD, he2020pathvqa, gurari2018vizwiz, winata2024worldcuisines, masry2022chartqa, chou2020VQA360, hsu2022geoclidean}. Many also incorporate multimodal challenges across domains.

Specialized datasets are developed for specific domains and applications, including medical imaging (\eg, VQA-RAD~\citep{lau2018VQARAD}, PathVQA~\citep{he2020pathvqa}), visually impaired assistance (\eg, VizWiz~\citep{gurari2018vizwiz}), multilingual and cultural diversity (\eg, WorldCuisines~\citep{winata2024worldcuisines}), mathematical reasoning (\eg, MathVista~\citep{lu2023mathvista}, ChartQA~\citep{masry2022chartqa}, Geoclidean~\citep{hsu2022geoclidean}), and embodied environments (\eg, VQA 360°~\citep{chou2020VQA360}). Additionally, comprehensive benchmarks such as MM-Vet~\citep{mmvetyu2024}, MME~\citep{fu2024mmebenchmark}, MMMU~\citep{yue2024mmmu}, and SEED-Bench~\citep{li2023seedbench} systematically evaluate vision-language capabilities across multiple reasoning domains. Benchmarks including MMBench~\citep{liu2024mmbench}, LLaVA-Bench~\citep{liu2023improvedllavabench}, NovPhy~\citep{pinto2024novphy}, M$^3$CoT~\citep{chen2024m3cot}, and VisIT-Bench~\citep{bitton2023visit} further probe fine-grained skills such as instruction-following, physical reasoning, and visual conversational interactions.

\textbf{Hallucination and Robustness Evaluation.}
This category includes benchmarks designed to detect hallucinated outputs, breakdowns in robustness, and model inconsistency, particularly in cases where the alignment between vision and language inputs is unreliable or weak. These tasks typically involve verifying whether model-generated responses are faithfully grounded in the input image and whether the model maintains consistent behavior under various perturbations.
Representative datasets such as POPE~\citep{li2023pope}, HallusionBench~\citep{guan2024hallusionbench}, YESBUT~\citep{liang2025yesbut}, and CausalVQA~\citep{Agarwal2019causalvqa} are designed to ensure that model outputs remain factually supported by visual evidence.

\subsection{Evaluation Metrics}
Evaluating visual reasoning involves a diverse set of metrics designed to capture not only the correctness of answers but also the interpretability, grounding consistency, and efficiency of the reasoning process. These metrics are typically selected based on task characteristics, dataset design, and the specific reasoning abilities being assessed.

\textbf{Accuracy-Based Evaluation.} Accuracy and exact match remain the dominant metrics for classification-based tasks, including general VQA, compositional reasoning (\eg, GQA~\citep{hudson2019gqa}), and knowledge-intensive VQA (\eg, OK-VQA~\citep{marino2019okvqa}, A-OKVQA~\citep{schwenk2022aokvqa}). For tasks with variable answers, such as TextVQA~\citep{kovaleva2020textvqa} and ST-VQA~\citep{biten2019STVQA}, metrics like VQA Score are used. A common evaluation paradigm in these benchmarks is option matching, a closed-set method where models select from predefined candidates. This reduces ambiguity in open-ended tasks and ensures standardized evaluation. Datasets such as CLEVR~\citep{johnson2017clevr}, ScienceQA~\citep{lu2022scienceqa}, MME~\citep{fu2024mmebenchmark}, and MMBench~\citep{liu2024mmbench} rely on this approach to improve fairness and comparability.

\textbf{Grounding and Segmentation Metrics.} For tasks requiring visual localization, metrics such as Intersection over Union (IoU), and generalized IoU~\citep{rezatofighi2019generalized} are widely used to assess the alignment between predicted and ground-truth regions. Metrics like AP@50 and AP@75 evaluate precision at different IoU thresholds. Tasks involving visual reasoning with segmentation leverage grounding accuracy and salient object detection metrics, including S-measure and mean absolute error.

\textbf{Rationale and Text Quality.} To evaluate explanation quality, metrics such as BLEU~\citep{papinenietal2002bleu}, ROUGE~\citep{lin2004rouge}, and CLIP sentence similarity~\citep{hessel-etal-2021-clipscore} are used to assess the coherence and informativeness of generated rationales. F1, precision, and recall are employed for sequential tasks or when answers involve structured text. Metrics like answer-grounding consistency and average output token length further probe whether generated explanations align with visual evidence and maintain concise reasoning.

\textbf{Task-Specific and Behavioral Metrics.} Visual reasoning systems are further evaluated using specialized metrics such as search length (number of steps in visual search)~\citep{wu2024vstar}, action counts (frequency of operations like zoom or segmentation)~\citep{zhou2024imageofthought}, hallucination rate (frequency of unsupported content)~\citep{OpenAI-o3-o4mini-2025}, and confidence improvement (sensitivity to question variations)~\citep{kim2025visualsurvey}. Keypoint-aware evaluation and format adherence metrics are employed in manipulation and program-based reasoning settings.

\textbf{Efficiency and Cost.} Finally, metrics related to resource usage, such as runtime, inference latency, LLM token consumption, and financial cost, are reported to assess the computational efficiency and scalability of vision-language models~\citep{cai2025naver}.

These evaluation metrics collectively provide a comprehensive view of model performance, capturing both end-task accuracy and the underlying reasoning behavior, interpretability, and resource efficiency necessary for reliable multimodal intelligence.

\subsection{Synthesis and Outlook}
Current benchmarks and evaluation protocols provide a broad landscape for assessing visual reasoning, offering evaluations across final answer accuracy, visual grounding, textual explanation quality, and domain coverage. They support tasks ranging from general VQA to fine-grained perception, knowledge-based reasoning, and multimodal interactions, spanning diverse modalities, task types, and reasoning skills. Despite this breadth, several critical limitations remain.

An important limitation in current evaluation practices is that existing benchmarks rarely evaluate explicit intermediate reasoning steps, also referred to as step-level or thought-step outputs, which have become increasingly central in recent compositional visual reasoning methods. Traditional evaluation metrics like BLEU~\citep{papinenietal2002bleu}, ROUGE~\citep{lin2004rouge}, and grounding scores provide only indirect assessments by evaluating final rationales or explanations. Consequently, they fail to directly measure the accuracy, coherence, or causal consistency of individual reasoning steps. For instance, a model might produce correct final answers but rely on incorrect intermediate deductions or superficial correlations.

Second, current benchmarks lack explicit quantification and analysis of reasoning difficulty levels, making it challenging to understand the distribution of question complexity. Questions involving a single object and its attributes are significantly easier compared to those involving multiple objects with intricate attribute references and complex inter-object relationships. For example, a question like ``What color is the ball?'' demands far simpler reasoning than ``Which object next to the small blue cube is the largest red sphere?'' Without explicitly defined difficulty scales, researchers cannot accurately evaluate a model's capacity to handle varying complexity levels.

To address these issues, future benchmarks should: 1) incorporate detailed annotations and corresponding metrics to evaluate intermediate reasoning steps, with a particular focus on assessing step-level consistency, causal coherence, and grounding accuracy; 2) explicitly quantify and categorize reasoning difficulty levels, enabling systematic evaluation of model performance across a spectrum of question complexities; and 3) provide comprehensive diagnostic insights by clearly differentiating reasoning capabilities, thereby helping researchers identify specific limitations in models and informing targeted improvements. Advancing these directions will enable deeper insights into model behaviors and support the robust development of compositional visual reasoning frameworks.

\section{Insights, Challenges and Directions}
\label{sec:challenge}

Compositional visual reasoning (CVR) has evolved quickly in recent years, with numerous papers emerging across different classes of models and evaluated on a wide range of benchmarks. In this section, we synthesize key insights from recent research, identifying trends and promising principles. We then investigate core challenges that continue to limit scalability, generalization, and robustness, as well as examine future directions aimed at overcoming these challenges to advance general-purpose CVR systems.

\subsection{Insights}
Recent studies in compositional visual reasoning indicate a notable shift from static, perception-based approaches toward structured, multi-step reasoning frameworks inspired by cognitive science. Models such as Chain-of-Manipulations~\citep{qi2024cogcom}, and Visual CoT Prompting~\citep{chen2024vctp} introduce modular reasoning procedures that enable intermediate decision making and enhance interpretability. Dual-system frameworks, exemplified by FAST~\citep{sunfast}, emulate both intuitive and deliberative cognitive processes, while reinforcement learning–based methods like Ground-R1~\citep{cao2025groundr1} and Vision-R1~\citep{huang2025visionr1} leverage feedback signals to promote self-corrective and generalizable reasoning without heavy reliance on supervised annotations.

This trend is further exemplified by the emergence of agentic VLMs that integrate control flow, memory, and visual grounding into a unified reasoning loop. Architectures such as SEAL~\citep{wu2024vstar} and ZoomEye~\citep{shen2024zoomeye} enable iterative, goal-directed visual exploration guided by questions, while Griffon-R~\citep{zhan2025griffonr} fuses perception and reasoning within a single forward pass. These models demonstrate the potential of embedding planning, action selection, and evidence integration within an agentic framework for compositional tasks.

To address data scarcity, researchers have turned to automated pipelines that synthesize supervision using LLMs and vision models. These pipelines generate CoT rationales, visual masks, and object-level annotations, providing rich multimodal supervision. Benchmarks such as MM-GCoT~\citep{Wu_2026_CVPR_mmgcot} and VoCoT~\citep{li-etal-2025-vocot} emphasize step-wise reasoning and grounded verification, contributing to more interpretable and robust evaluation protocols.

At the architectural level, ongoing efforts aim to unify static scene understanding with dynamic interaction and planning capabilities. This includes the development of tool-aware LLMs for adaptive tool invocation and coordination, as well as models with visual working memory mechanisms that manage long-horizon reasoning over fine-grained visual inputs. Models like DC2~\citep{wang2025dc2} and SEAL~\citep{wu2024vstar} illustrate the effectiveness of such memory-based designs in maintaining contextual coherence while minimizing computational overhead.

\subsection{Key Challenges and Future Directions}
Despite the rapid progress in compositional visual reasoning, significant challenges remain in enhancing reasoning capability, improving data quality, integrating architectures, and refining evaluation methodology. Addressing these limitations is essential for developing scalable, trustworthy, and general-purpose compositional visual reasoning systems.

\subsubsection{Limitations of LLM-Based Reasoning}

A central limitation across all five stages of compositional visual reasoning systems discussed is their heavy reliance on LLMs as the core reasoning module. While LLMs excel at processing natural language and generating coherent text, they often lack the grounded, structured reasoning necessary for complex visual tasks. In particular, LLMs lack an internal world model, which limits their ability to simulate visual or physical dynamics, perform spatial transformations, or reason about hypothetical or counterfactual outcomes. These capabilities are essential for tasks such as object reconfiguration, spatial prediction, and planning tool usage~\citep{hao2023reasoning}.

For example, when tasked with predicting whether one object can fit into another or reasoning about the consequence of moving an obstacle, LLMs tend to rely on linguistic priors rather than perceptual evidence, often producing plausible but incorrect answers. This limitation is rooted in their autoregressive architecture, which is optimized for next-token prediction rather than iterative self-correction or causal inference~\citep{kambhampati2024cant_plan}. 

From a cognitive standpoint, this behavior resembles System 1 reasoning—fast, heuristic, and shallow—rather than System 2 reasoning, which involves deliberate, multi-step analysis grounded in perceptual input and symbolic manipulation~\citep{huang2025planning, sloman1996two_system, wangpromptagent}. This gap is especially evident in visual reasoning benchmarks that require multi-hop spatial inference, counterfactual exploration, or tracking of visual state changes, where LLMs often fail to reason beyond surface-level correlations~\citep{yehudai2025survey, li2024lasp}.

\textbf{Potential Directions.} To address these challenges, one promising solution is to introduce an explicit internal world model into compositional visual reasoning systems. Such models would enable agents to simulate hypothetical visual scenarios, reason over spatial and temporal dynamics, and plan multi-step actions grounded in perception. This capability supports forward simulation, counterfactual inference, and long-horizon planning. It also aligns more closely with System 2 reasoning. Incorporating world models takes inspiration from model-based reinforcement learning and offers a pathway to bridging the gap between language-driven reasoning and perceptually grounded intelligence~\citep{cao2025groundr1}.

\subsubsection{Hallucinations}
Although LLMs are highly effective at language-based inference, their reasoning often lacks a strong connection to visual content, particularly when visual inputs are projected into language space without sufficient grounding. This misalignment can lead to hallucinated conclusions, superficial associations, and inaccurate spatial or relational reasoning~\citep{zhou2024imageofthought, chen2024p2g}. For instance, when presented with an image of a green banana, an LLM may incorrectly respond that the banana is yellow, reflecting linguistic bias rather than perceptual evidence.

Compositional approaches, while modular in structure, are not fully immune to hallucination and unfaithfulness. Models can produce plausible yet unsupported intermediate steps or final answers, especially when reasoning components operate on incomplete or poorly grounded visual information~\citep{qi2024cogcom, zhan2025griffonr, ke2025dwim}. These issues may arise from semantic mismatches between visual inputs and symbolic reasoning, limited supervision during grounding, or an overdependence on prior knowledge encoded in the LLM~\citep{khan2025_neurosymbolic_VR_survey, kuang2025_natural_language_vqa_survey}.

Furthermore, shortcut learning persists even in multi-step pipelines, where models exploit spurious correlations within decomposed tasks rather than reason from visual evidence. These issues limit generalization, decrease interpretability, and reduce confidence in system outputs, highlighting the need for improved alignment, step-level supervision, and stronger grounding mechanisms in compositional reasoning frameworks~\citep{Wu_2026_CVPR_mmgcot, sunfast}.

\textbf{Potential Directions.} Given the ambiguity and context-dependence of real-world CVR tasks, incorporating interactive frameworks with human-in-the-loop supervision holds promise. Allowing users to guide, verify, or revise intermediate reasoning steps can enhance both system reliability and user trust. Such interaction also supports adaptive learning by providing feedback signals that reinforce faithful reasoning.

\subsubsection{Bias Toward Deductive Reasoning}

Most current compositional visual reasoning systems predominantly rely on deductive reasoning, where inference steps follow formal logic. However, this approach assumes the correctness of initial premises and can produce erroneous conclusions when the inputs are noisy or biased~\citep{johnson1999deductive, wang2024reasoning_mllm_survey}. For example, a system might conclude that an object is a chair because it has four legs and a backrest, even if the object is actually a sculpture, due to flawed visual grounding. 

\textbf{Potential Directions.} To improve robustness and adaptability, alternative reasoning paradigms offer promising directions. Inductive reasoning enables models to generalize patterns from visual observations—such as learning affordances from multiple object instances~\citep{weston2015towards, yu2024natural}. Abductive reasoning supports generating plausible explanations when visual input is ambiguous, for instance, inferring that a spilled drink on the floor suggests a tipped-over glass nearby~\citep{prabhushankar2022introspective, Yu2021ASO}. Analogical reasoning facilitates relational transfer, such as applying knowledge about assembling furniture from diagrams to configuring unfamiliar mechanical parts in a robotics task, where similar spatial relations and assembly logic are required~\citep{crouse2021neural, Blokpoel2019DeepAI}.

\subsubsection{Data Limitations and Scalability Challenges}
Progress in compositional visual reasoning is significantly hindered by data-related constraints. Existing datasets lack sufficient compositional diversity, multi-step supervision, and grounded visual programs, limiting the development of robust reasoning capabilities~\citep{patel2024tripletclip, wu2025compact}. High-quality annotations for step-by-step inference are costly and difficult to scale~\citep{chen2024p2g, huang2025visionr1, zhang2024provision}, while synthetic pipelines introduce noise due to unreliable perception tools~\citep{ke2025dwim, gao2024clova}.

Supervised fine-tuning on curated CoT data can cause overfitting and cognitive rigidity, restricting generalization~\citep{hu2024pali, tan2025reasonrft, ke2025dwim}. Automatically generated instruction data often lacks reasoning depth and diversity, and even advanced models like GPT-4V can yield unfaithful or inconsistent outputs~\citep{qi2024cogcom, sunfast}. Benchmarks such as CLEVR~\citep{johnson2017clevr} and VQA~\citep{antol2015vqa} still emphasize shallow tasks, overlooking multi-step reasoning~\citep{Wu_2026_CVPR_mmgcot, jahangard2025jrdb}.

\textbf{Potential Directions.} Addressing data scarcity and scalability challenges requires the development of more efficient and grounded data construction strategies. One promising direction is to build hybrid data engines that combine synthetic scene generation with real-world imagery, augmented by perception-enhanced simulation environments~\citep{le2024jrdb, ehsanpour2022jrdb}. These platforms can generate richly annotated multi-step reasoning traces with minimal manual effort. Additionally, leveraging weak supervision, self-training, and feedback-driven refinement can mitigate annotation costs while improving fidelity and reasoning diversity. Instruction tuning with human-in-the-loop filtering or discrepancy-aware self-verification can further enhance data quality~\citep{khan2024visrep, ke2025dwim, ma2024mmtooluse}.

\subsubsection{Tool Integration and Architectural Bottlenecks}
Compositional visual reasoning faces architectural challenges in the integration of LLMs with vision modules and external tools. Current systems often lack tool awareness, incur high computational costs during tool use, and struggle with re-planning based on intermediate results~\citep{ke2024hydra, ke2025dwim, lu2023chameleon, khan2024visrep}. Vision encoders introduce information bottlenecks by projecting low-resolution inputs into limited token spaces~\citep{wu2024vstar, radford2021clip, ke2025dwim}, while fragmented pipelines that treat static and dynamic content separately weaken cross-modal grounding~\citep{gao2024clova, hu2024pali, liu2024llavaplus, fei2024vitron}.

Tool learning is hindered by retrieval constraints, inconsistent tool formats, limited generalization to unseen tools, and underdeveloped multimodal capabilities~\citep{gao2023assistgpt, lu2023chameleon}. Balancing reasoning depth with efficiency remains an open issue~\citep{xi2025lightweight}.

\textbf{Potential Directions.} Future research should explore more integrated architectures that support tighter coupling between perception, reasoning, and tool execution. Promising directions include improving tool-awareness, designing efficient planning and re-planning strategies, enhancing cross-modal grounding, and supporting scalable generalization to unseen tools.

\subsubsection{Benchmark Limitations and Evaluation Contamination}
Current benchmarks often overestimate model capabilities by relying on in-distribution data and biased annotations~\citep{hsieh2023sugarcrepe, abbasi2024deciphering}. Models frequently exploit linguistic shortcuts instead of engaging in faithful visual reasoning~\citep{kim2025visualsurvey, ma2024mmtooluse}. Evaluation contamination arises from synthetic queries, artificial distribution shifts, and browsing-enabled LLMs, which obscure reasoning failures~\citep{OpenAI-o3-o4mini-2025, yang2023mmreact}. 

A particularly critical gap lies in the evaluation of compositional visual reasoning models. Although these methods emphasize explicit intermediate reasoning steps, current benchmarks typically assess performance based solely on final answers, without evaluating the quality or faithfulness of the intermediate reasoning process. This evaluation limitation undermines interpretability and hampers progress toward robust, generalizable visual reasoning~\citep{wu2025visco}.

\textbf{Potential Directions.} Future progress relies on the development of more nuanced evaluation protocols that move beyond final answer accuracy. Promising directions include difficulty-aware scoring, assessment of intermediate reasoning steps, and multimodal explanation evaluation. Moreover, extending benchmarks to include richer visual compositions, multi-hop reasoning trajectories, and step-level annotations will be crucial for advancing robust and generalizable compositional visual reasoning~\citep{ma2024robustsurvey}.

\section{Conclusion}
\label{sec:conclusion}
In this survey, we focus on recent advances in compositional visual reasoning (CVR), a rapidly emerging research direction that emphasizes structured, multi-step reasoning over visual inputs. Unlike monolithic vision-language models that directly map input to output in a single pass, CVR frameworks aim to explicitly bridge perception and reasoning through intermediate, interpretable steps. We highlight the key distinctions between these two paradigms, illustrating why compositional reasoning is essential for achieving robust and cognitively aligned intelligence.


This survey is organized around four core questions: Why is compositional visual reasoning necessary? What are the major architectural shifts and reasoning paradigms in this field? What benchmarks and evaluation protocols are currently used? What are the prevailing limitations and open challenges?
To answer these questions, we reviewed over $100$ papers and distilled the field into five major stages of development: prompt-enhanced language-centric methods, tool-enhanced LLMs, tool-enhanced VLMs, chain-of-thought VLMs, and unified agentic VLM-based frameworks. For each stage, we introduce representative works, analyze their design principles, and summarize their strengths and limitations. We also provide a roadmap that charts the evolution of CVR systems and offers insights for future research.
In addition, we review benchmark datasets and metrics, and highlight key challenges such as visual grounding, hallucination, data scalability, system bottlenecks, benchmark limits, and evaluation contamination. 


\section*{ACKNOWLEDGMENT}
This work was supported by Building 4.0 CRC and the Commonwealth of Australia through the Cooperative Research Centres Program. It was also partially funded by the DARPA ANSR program (FA8750-23-2-1016) and the ARC DECRA program (DE250100032).

\clearpage
{
    \small
    \bibliographystyle{unsrt}
    \bibliography{main}
}


\end{document}